\documentclass[10pt,twocolumn,letterpaper]{article}

\usepackage{cvpr}              % To produce the CAMERA-READY version
\usepackage{makecell}
\usepackage{multirow}
\usepackage{graphicx}
\usepackage{enumitem}
\usepackage[table]{xcolor}
\usepackage{soul, color, xcolor}
\usepackage{tabularray}
\usepackage{pifont}       % \ding{xx}
\usepackage{bbding}       % \Checkmark,\XSolid,... (需要和pifont宏包共同使用)
\usepackage{fontawesome}  % \faCheck,\faTime
\usepackage{tabularx}

\usepackage{marvosym}
\usepackage[accsupp]{axessibility}
\usepackage{tikz}
\usepackage{ragged2e}

\newcommand{\circnum}[1]{%
\tikz[baseline=(char.base)]{
    \node[shape=circle, fill=black, inner sep=1pt] (char)
    {\color{white}\small #1};
}}

\newif\ifshowcomments
\showcommentstrue

\ifshowcomments
    \newcommand{\xhwu}[1]{\textcolor{teal}{[XHWU: #1]}}
    \newcommand{\todo}[1]{\textcolor{magenta}{[TODO: #1]}}
\else
    \newcommand{\xhwu}[1]{}
    \newcommand{\todo}[1]{}
\fi

\newcommand{\cmark}{\ding{51}} % ✓
\newcommand{\xmark}{\ding{55}} % ✗

\renewcommand{\todo}[1]{{\color{red}#1}}

\definecolor{cvprblue}{rgb}{0.21,0.49,0.74}
\usepackage[pagebackref,breaklinks,colorlinks,allcolors=cvprblue]{hyperref}

\def\confName{CVPR}
\def\confYear{2026}

\def\Method{CREval}
\title{\Method: An Automated Interpretable Evaluation for Creative Image Manipulation under Complex Instructions}

\author{Chonghuinan Wang$^{1}$ \quad Zihan Chen$^1$ \quad Yuxiang Wei$^1$ \quad Tianyi Jiang$^1$ \\ Xiaohe Wu$^1$\textsuperscript{\Letter} \quad Fan Li$^2$\textsuperscript{\dag} \quad Wangmeng Zuo$^{1,3}$ \quad Hongxun Yao$^1$ \\
$^1$Harbin Institute of Technology \quad $^2$Huawei Noah’s Ark Lab \quad $^3$Pengcheng Lab, Guangzhou\\
{\tt\small \{25b903050, zhchen\}@stu.hit.edu.cn},
{\tt\small \{yuxiang.wei.cs, 1643026263jty, csxhwu\}@gmail.com} \\
{\tt\small lifan61@huawei.com},
{\tt\small \{wmzuo, h.yao\}@hit.edu.cn} \\
{\tt\small \textcolor{blue}{https://github.com/ChonghuinanWang/CREval}}
}
\begin{document}
\twocolumn[{
    \renewcommand\twocolumn[1][]{#1}
    \maketitle
    \begin{center}
        \captionsetup{type=figure}
        \includegraphics[width=\textwidth]{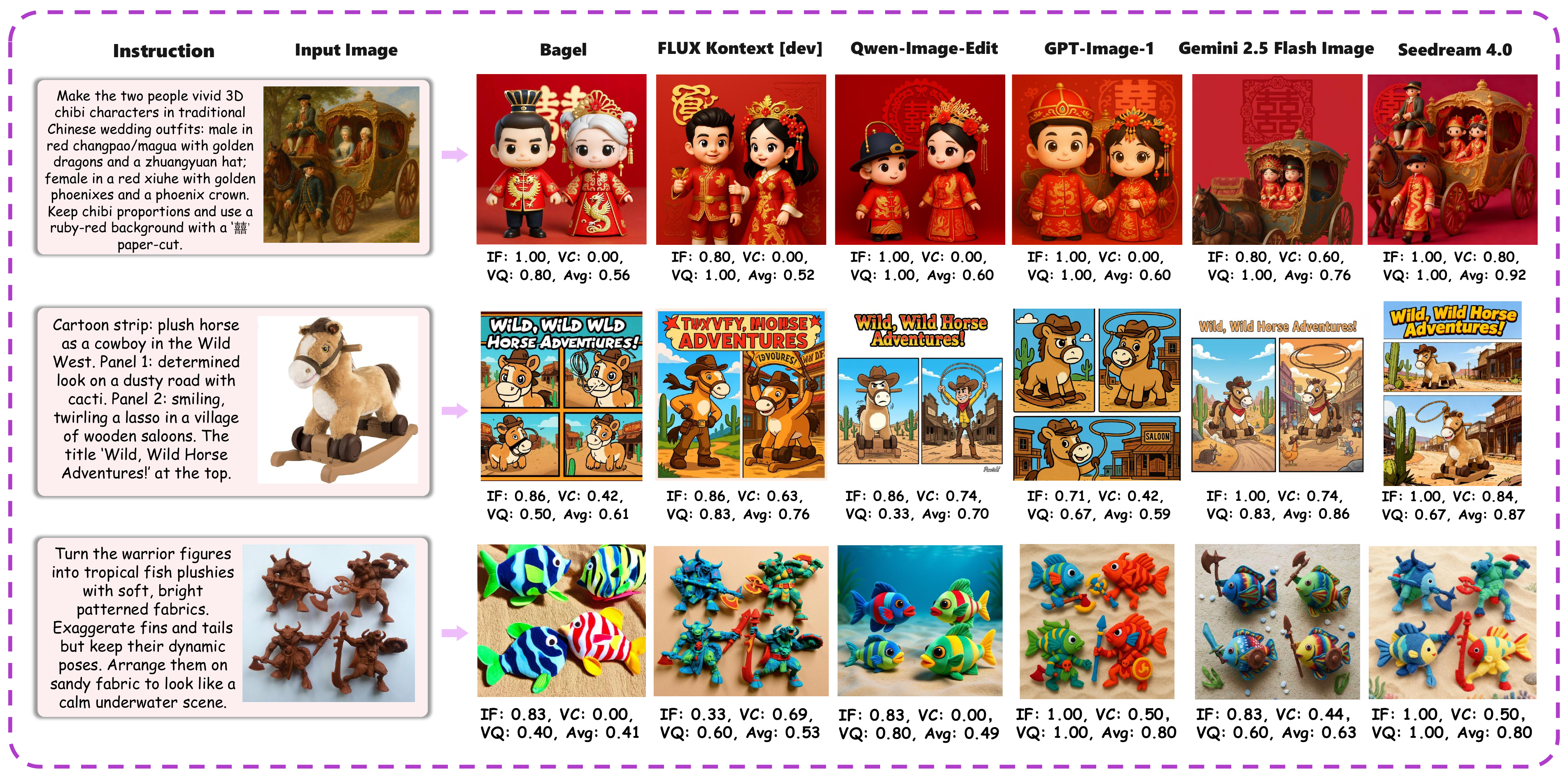}
        \captionof{figure}{\textbf{Evaluation of state-of-the-art image generation and editing models using CREval, with GPT-4o serving as the evaluator.} Each edited image is evaluated across three metrics: Instruction Following (IF), Visual Consistency (VC), and Visual Quality (VQ). The results indicate that the complex and creative instructions in CREval-Bench pose substantial challenges for current image manipulation models.}
        \label{fig:teaser}
    \end{center}
}]

\renewcommand{\thefootnote}{}
\footnotetext{\Letter \space Corresponding Author, \space \dag \space Project Leader}

\begin{abstract}
% Instruction-based multimodal image editing has recently achieved rapid progress. However, existing evaluation protocols lack a systematic and standardized framework for assessing model performance under complex and creatively challenging instructions. To address this limitation, we propose CREval-Bench, which is the first comprehensive benchmark specifically designed for creative image editing under complex instruction. CREval-Bench covers 3 categories and 9 creative dimensions, encompassing over 800 editing samples and 13K evaluation queries. We further develop CREval, a fully automated, multi-dimensional evaluation pipeline, substantially reducing the cost and subjectivity of manual assessments. 
% We benchmark a range of state-of-the-art open-source and closed-source models, revealing that while the closed-source models generally outperform open-source systems on complex creative editing tasks, all models exhibit substantial room for improvement. Furthermore, human similarity studies demonstrate strong alignment between our automated metrics and human judgments. 
% Overall, CREval establishes a scalable and reliable foundation for evaluating creative image editing models and highlights key challenges and opportunities for future research. Code and data will be released publicly.

%-lifan 1108
Instruction-based multimodal image manipulation has recently made rapid progress. However, existing evaluation methods lack a systematic and human-aligned framework for assessing model performance on complex and creative editing tasks. 
To address this gap, we propose CREval, a fully automated question–answer (QA)–based evaluation pipeline that overcomes the incompleteness and poor interpretability of opaque Multimodal Large Language Models (MLLMs) scoring. Simultaneously, we introduce CREval-Bench, a comprehensive benchmark specifically designed for creative image manipulation under complex instructions. CREval-Bench covers three categories and nine creative dimensions, comprising over 800 editing samples and 13K evaluation queries.
Leveraging this pipeline and benchmark, we systematically evaluate a diverse set of state-of-the-art open and closed-source models. The results reveal that while closed-source models generally outperform open-source ones on complex and creative tasks, all models still struggle to complete such edits effectively. In addition, user studies demonstrate strong consistency between CREval's automated metrics and human judgments.
Therefore, CREval provides a reliable foundation for evaluating image editing models on complex and creative image manipulation tasks, and highlights key challenges and opportunities for future research. 
% All code and data will be released publicly.
% All code and data are publicly available at https://github.com/ChonghuinanWang/CREval.
\end{abstract}

\begin{figure*}[h!]
    \centering
    \includegraphics[width=\textwidth]{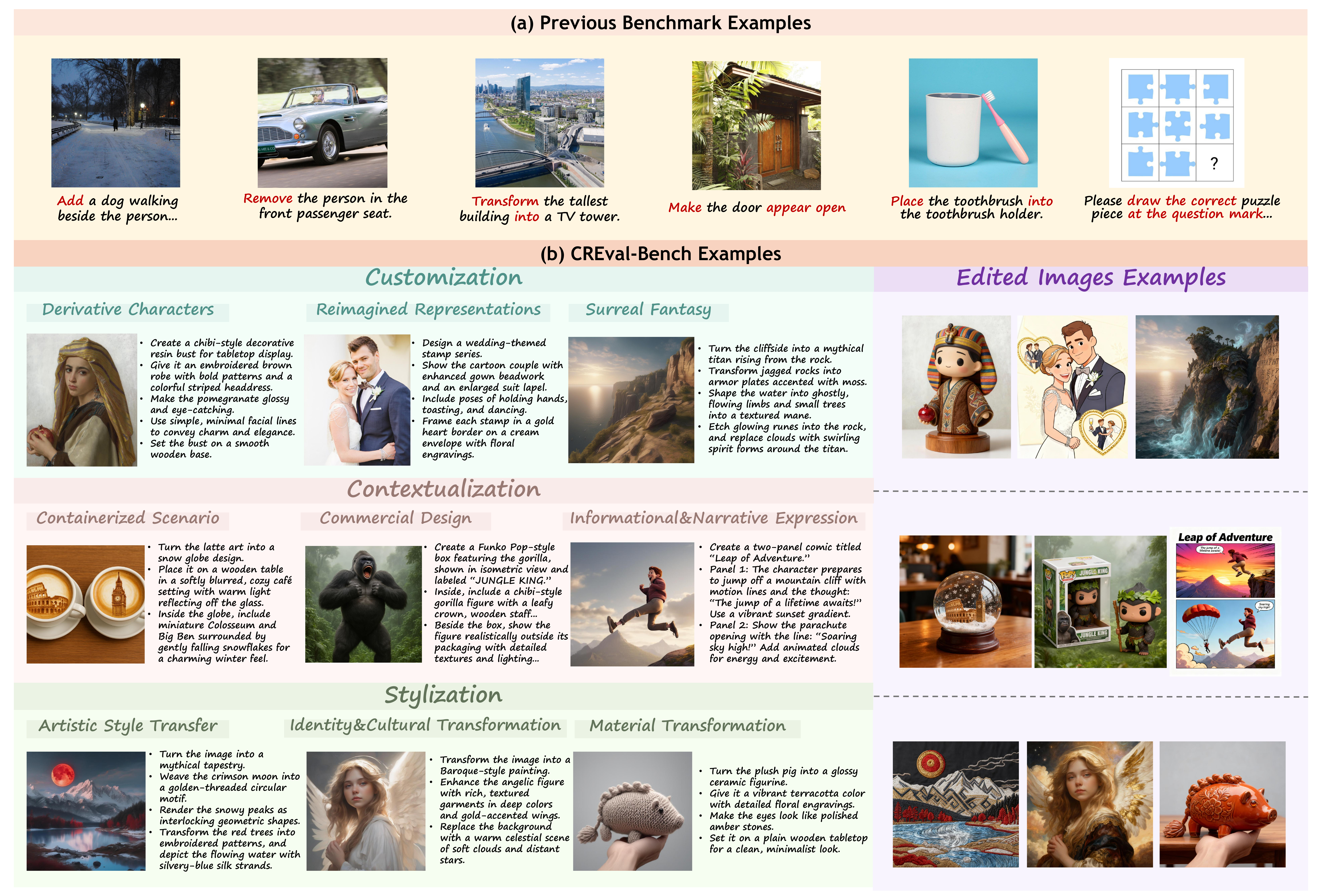}
    \caption{\textbf{Comparison with previous benchmark.} The CREval-Bench dataset extends existing instruction-based editing benchmarks by incorporating more complex, creative, and semantically rich instructions. Such design facilitates a comprehensive evaluation of model performance in handling imaginative and complex instruction editing tasks. In (b), the edited image examples on the right correspond one-to-one with the image-instruction pairs on the left.}
    \label{fig:compare}
\end{figure*}

\section{Introduction}
\label{sec:intro}

In the context of image manipulation tasks guided by instructions, can models maintain robust performance when these instructions exhibit high complexity and innovative characteristics?

Currently, multimodal generative models~\cite{cao2025hunyuanimage30technicalreport, xu2025insightedit, zhou2025fireedit, qin2025unicotunifiedchainofthoughtreasoning, li2025manzanosimplescalableunified} have demonstrated remarkable capabilities in instruction-based image editing tasks. Notably, models like GPT-Image-1~\cite{openai2025gptimage1} and Gemini 2.5 Flash Image~\cite{google2025gemini25flash} have significantly improved their comprehension of complex instructions compared to earlier models~\cite{brooks2023instructpix2pix, ruiz2023dreambooth, shi2024seededit, sheynin2024emu, xiao2025omnigen1}.

However, current generative image generation and editing models still face significant challenges when handling complex instruction-based tasks, particularly in ``free-style creative image editing'' scenarios as illustrated in Figure~\ref{fig:teaser}. These challenges include: \circnum{1} Insufficient instruction following: Models struggle to accurately interpret and execute complex user instructions, leading to incomplete or incorrect edits; \circnum{2} Visual feature inconsistency: Models fail to preserve key visual characteristics of the subject's identity, resulting in a loss of core information; \circnum{3} Poor visual quality: Generated images often contain artifacts and distortions, diminishing reality and fidelity.
More critically, current evaluation benchmarks~\cite{basu2023editval, hui2024hq, NEURIPS2024_48fecef4, ye2025imgedit, luo2025editscore, chen2025edival, wu2025kris, sun2025t2i} primarily focus on common tasks like object addition, replacement, deletion, color adjustment, and logical reasoning as shown in Figure~\ref{fig:compare} (a). These works fail to effectively evaluate the performance of free-style creative image editing tasks, as illustrated by the example case in Figure~\ref{fig:compare} (b).
%-end

% Addressing this core gap, this paper proposes and constructs CREval-Bench. Unlike previous benchmarks, as shown in Figure~\ref{fig:compare}, this is an evaluation framework specifically designed for complex instruction-based creative image editing tasks. CREval-Bench employs a fully automated evaluation process to ensure objectivity and reproducibility, covering three major categories of creative editing tasks and subdivided into nine specific evaluation dimensions. 
% In terms of data scale, CREval-Bench comprises over 13K evaluation samples. Its assessment process uses the GPT-4o model, taking evaluation quadruplets consisting of ${[I_i, I_o, P, Q]}$, where ${I_i}$ denotes original input image, ${I_o}$ represents edited output image, ${P}$ is editing instruction, and ${Q}$ signifies question, as input to compute quantitative scores, ultimately producing quantifiable evaluation results.

\begin{table}[t]
    \centering
    \caption{\textbf{Comparison to other existing benchmark.} Our benchmark provides a comprehensive evaluation of creative image manipulation by leveraging VQA-based scoring.}
    \resizebox{\linewidth}{!}{
    
        \begin{tabular}{ccccccc}
             \noalign{\hrule height 1.5pt}
             Dataset & Size & Scoring & Creative & Fully-Automatic \\
             \hline
             
             ImgEdit-Bench ~\cite{ye2025imgedit} & 791 & MLLMs scoring & \xmark & \cmark \\
             % VIEScore &  & MLLM scoring & \xmark & \cmark \\
             % I2EBench & 2K & VQA scoring & \xmark & \ \cmark  \\
             KRIS-Bench ~\cite{wu2025kris} & 1267 &  MLLMs scoring & \xmark & \cmark \\
             RISE-Bench ~\cite{zhao2025envisioning} & 360 & MLLMs scoring & \xmark & \cmark \\
             GEdit-Bench ~\cite{liu2025step1x-edit} & 606 & MLLMs scoring & \xmark & \cmark \\ 
             \rowcolor{gray!20} CREval-Bench & 874 & VQA scoring & \cmark & \cmark \\
             \noalign{\hrule height 1.5pt}
             
        \end{tabular}
    }\label{tab:compare_with_previous_benchmark}
\end{table}
%-lifan-start
To address this gap, we propose CREval, a fully automated, multidimensional evaluation pipeline, along with a benchmark named CREval-Bench. This framework is specifically designed to provide objective, fully automated evaluation for complex instruction–based creative image manipulation. CREval-Bench covers three major categories and nine creative dimensions, comprising over 800 editing samples and 13K evaluation queries. 
% \tocheck{
As summarized in Table ~\ref{tab:compare_with_previous_benchmark}, in contrast to prior approaches that rely solely on Multimodal Large Language Models (MLLMs) as holistic evaluators, we decompose the evaluation process into three complementary metrics: Instruction Following (IF), Visual Consistency (VC), and Visual Quality (VQ). For each sample, we derive targeted Question–Answer (QA) pairs grounded in the original image and the associated instruction. 
%
% Instead of having MLLMs directly assign scores, as this approach typically results in incomplete coverage of evaluation dimensions and poor interpretability with no transparency into the reasons for scoring, we prompt MLLMs to respond to these structured queries.
Instead of directly asking MLLMs to assign scores, which often leads to incomplete coverage of evaluation dimensions and limited interpretability due to the lack of transparent scoring rationales, we prompt MLLMs to respond to these structured queries.
This enables transparent scoring because the responses explicitly indicate where points should be awarded or deducted, leading to more comprehensive and interpretable evaluation.
% }
The evaluation process uses MLLMs (\eg, GPT-4o) to compute quantitative scores from triplets ${[I_i, I_o, Q]}$, which represent the input image, output image, and evaluation query. These triplets are fed into the MLLMs to generate objective evaluation results.
%-end

We conduct a comprehensive evaluation of mainstream open- and closed-source editing models using CREval. The results show that Seedream 4.0~\cite{seedream2025seedream} achieves the highest overall performance, exhibiting a strong balance across IF, VC, and VQ. Qwen-Image-Edit-2509~\cite{wu2025qwenimagetechnicalreport} ranks second, with both models outperforming GPT-Image-1~\cite{openai2025gptimage1} in aggregate performance. 
While closed-source models currently lead in overall performance, open-source models have shown promising progress. With continued technological advancement and community development, the competitiveness of open-source models is expected to further improve in the near future.

% \begin{itemize}
%     % \item We systematically categorized creative image editing systems based on complex instructions.
%     \item We introduce \textbf{CREval}, a fully automated \textbf{E}valuation framework for \textbf{C}\textbf{R}eative image editing under complex instruction, eliminates the need for manual human assessment. The framework is multidimensional and interpretable, addressing the limitations of traditional single-dimensional metrics that fail to capture comprehensive model performance. 
%     \item We build a \textbf{CREval-Bench} specifically designed to evaluate diverse creative editing scenarios, encompassing 3 categories and 9 dimensions, thereby establishing a foundation for systematic and fair assessment in this emerging research area.This a nolvelty benchmark for creative image edting under complex instructions.
%     \item We conduct extensive experiments on state-of-the-art image editing models. The results reveal both the strengths and limitations of these models in addressing flexible and creative editing tasks, providing valuable insights into current capabilities and future research directions.
%     \item Experiments demonstrates a strong correlation between CREval scores and human preference judgments, confirming its reliability and robustness.
% \end{itemize}

%--lifan-start
In summary, the main contributions of this paper are as follows:
\begin{itemize}
\item We propose \textbf{CREval}, a fully automated and QA-based evaluation framework for \textbf{C}\textbf{R}eative image manipulation under complex instructions, addressing the limitations of MLLMs scoring.
\item We build \textbf{CREval-Bench}, a comprehensive benchmark covering 3 categories and 9 dimensions to systematically and fairly evaluate diverse creative editing scenarios.
\item We conduct extensive experiments on state-of-the-art image generation and editing models, revealing their strengths and limitations in handling complex and flexible editing tasks, and providing insights for future research.
\item User studies demonstrate strong consistency between CREval scores and human preference judgments, confirming the reliability and robustness of the proposed evaluation framework.
\end{itemize}
%

%-------------------------------------------------------------------------

\section{Related Work}

\subsection{Instruction-based Image Editing Models}
Instruction-based image editing models aim to achieve semantic-level modifications and creations of image content by precisely understanding natural language instructions. The core challenge lies in balancing the accuracy of instruction adherence, the structural fidelity of the original image, and the generalization capability of the editing task.
Early works such as \cite{kawar2023imagic, brooks2023instructpix2pix, zhang2023magicbrush, hui2024hq, zhao2024ultraedit, li2024magiceraser, qin2025camedit} have achieved promising results in image editing guided by human instructions. ~\cite{NEURIPS2023_98530736_ImageBrush, zhao2024instructbrushlearningattentionbasedinstruction, sun2025pocketsr,wang2025ace} further extends this line of research by learning visual features to enable more precise and controllable image modifications. ~\cite{zhang2024instructeditinstructionbasedknowledgeediting, huang2024smartedit, kong2025dual} integrates MLLMs to enhance semantic understanding and reasoning capabilities, thereby addressing the limitations of CLIP-based text encoders, which often fail to produce satisfactory representations in complex scenarios.

Subsequent developments in image editing have shifted from single diffusion models toward multimodal Transformer~\cite{vaswani2017attention} and Flow Matching~\cite{lipman2022flow} architectures. Stable Diffusion 3~\cite{sd3esser2024scaling} replaces the conventional U-Net backbone with a multimodal diffusion Transformer (MMDiT) and adopts Flow Matching objectives for more stable and efficient training. ~\cite{hu2024instruct} introduces a unified framework for multimodal instruction-based editing, while ~\cite{kulikov2025flowedit} employs a pre-trained flow model to enable text-driven, non-reverse editing. Recent works such as ICEdit~\cite{zhang2025context}, Flux.1 Kontext~\cite{batifol2025flux} further demonstrate the advantages of Flow Matching architectures in enhancing instruction adherence and achieving fine-grained image modifications.

\subsection{Benchmarks for Image Editing}
The rapid development of instruction-based image editing models necessitates comprehensive evaluation frameworks to assess their capabilities systematically~\cite{zhang2023magicbrush, hu2023tifa, hui2024hq, zhao2024ultraedit, lin2024evaluating, NEURIPS2024_48fecef4}. To address data quality concerns and establish standardized protocols, MagicBrush~\cite{zhang2023magicbrush} introduced large-scale manual annotations for common object-level operations such as addition, removal, and replacement, 
% while subsequent efforts including HQ-Edit~\cite{hui2024hq}, UltraEdit~\cite{zhao2024ultraedit}, SEED-Data-Edit~\cite{ge2024seed}, AnyEdit~\cite{yu2025anyedit}, and ImgEdit-Bench~\cite{ye2025imgedit} scaled up synthetic pipelines with improved quality control and expanded editing taxonomies. 
while subsequent efforts including ~\cite{hui2024hq, zhao2024ultraedit, ge2024seed, yu2025anyedit, ye2025imgedit}

To enable more comprehensive and perceptually aligned evaluation,  I2EBench~\cite{NEURIPS2024_48fecef4} defines multiple evaluation dimensions spanning high-level semantics and low-level details, supported by extensive user studies for empirical validation. ~\cite{ku2023viescore, ryu2025towards, qian2025gie} proposed human-aligned evaluation protocols leveraging multimodal MLLMs and VQA-based functional correctness assessment.
These methods either partially rely on manual annotation or employ COCO-trained object detectors to identify and filter scene elements for automated evaluation, thereby constraining their applicability to more flexible and free-style editing scenarios.

Parallel efforts~\cite{jia2025compbench, wang2025complexbench, yang2025complexedit, li2025gir-bench, wu2025kris, zhao2025envisioning} targeting sophisticated scenarios evaluate multi-step reasoning or spatial complexity. However, these benchmarks primarily conceptualize complexity in terms of logical, compositional, or operational difficulty, without fully capturing the open-ended and creative challenges of free-style editing tasks.

% \url{https://www.computer.org/about/contact}.
\section{CREval-Bench}

\begin{figure}
    \centering
    \includegraphics[width=0.8\linewidth, trim=80 80 80 80,clip]{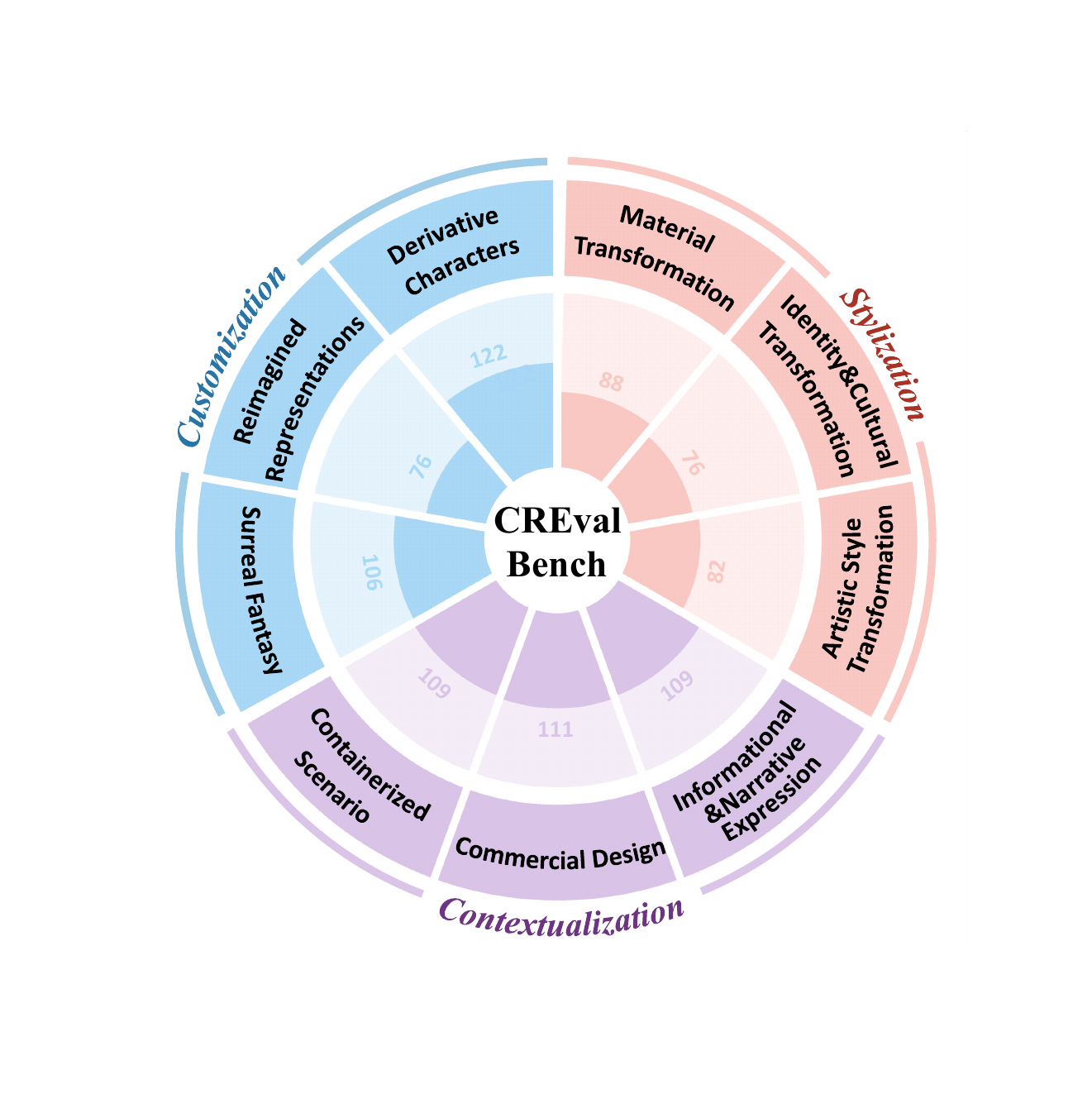}
    \caption{\textbf{Distribution of creative editing types.} Creative types are organized into 3 primary categories and 9 dimensions, with balanced sample counts to ensure comprehensive and consistent evaluation.}
    \label{fig:distribution}
\end{figure}

\begin{figure*}[h!]
    \centering
    \includegraphics[width=\textwidth]{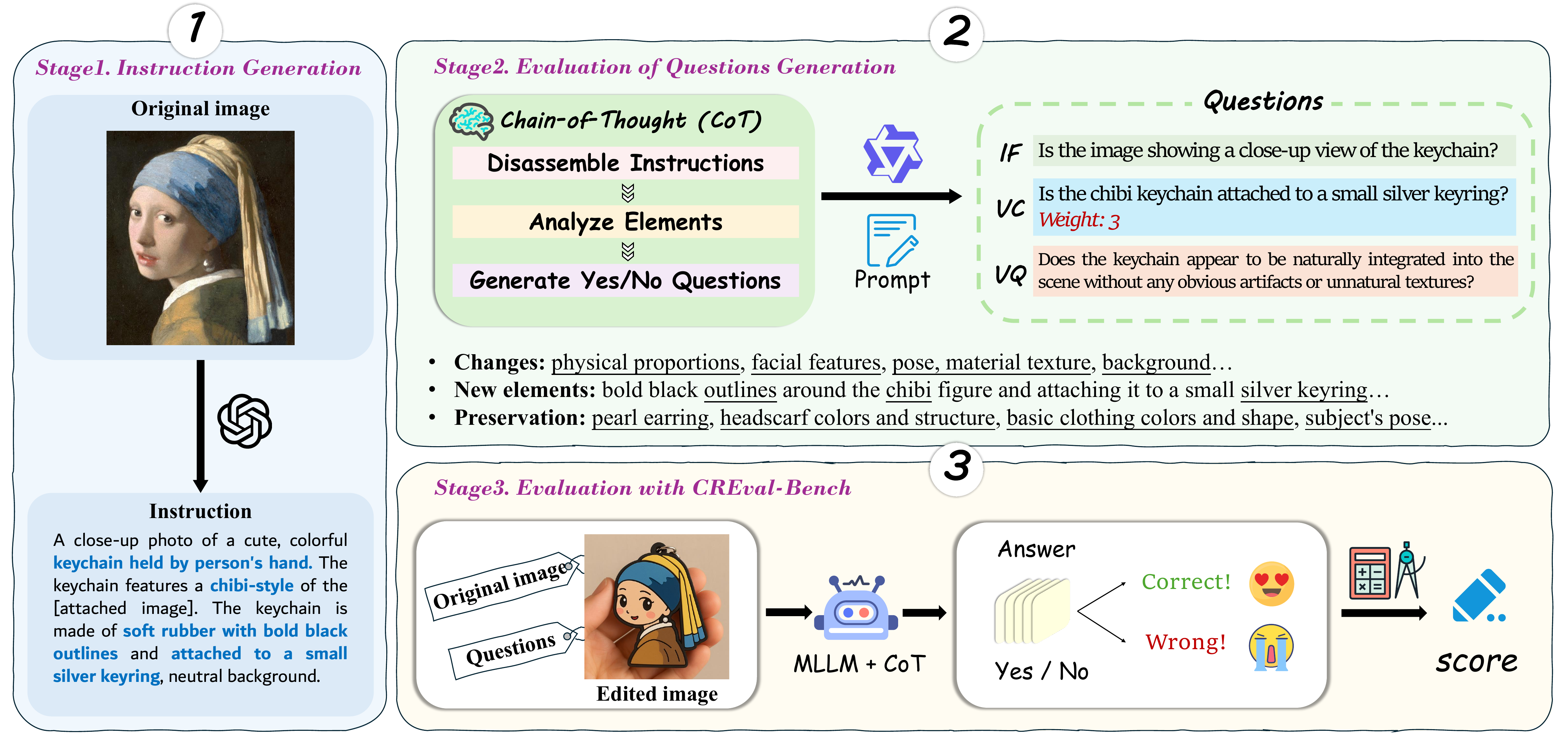}
    \caption{\textbf{Overview of CREval.} (1) In stage 1, we manually select high-quality images. We then construct several editing instruction examples and utilize the GPT-4o model for few-shot learning across 9 predefined dimensions, generating dimension consistent editing instructions and producing image–instruction pairs. (2) In stage 2, we use these image–instruction pairs to construct evaluation tasks. To reduce bias, we use different MLLMs such as Qwen2.5-VL-72B, to generate evaluation questions for 3 metrics using the Chain-of-Thought (CoT) method. Each metric contains at least 5 questions, with a total of no fewer than 15 questions per pair, completing the construction of the CREval-Bench. (3)In Stage 3, we evaluate mainstream image manipulation models using CREval method. A MLLM model is employed as the evaluator to score each edited image based on evaluation questions. The final performance metric is obtained by computing a weighted average score across all evaluation metrics.}
    \label{fig:global_step}
\end{figure*}

\label{sec:method}
% UniREditBench: A Unified Reasoning-based Image Editing Benchmark
% Creative image manipulation has become an important component of numerous visual computing applications, including advertising, artistic content creation, product design, and various everyday visual scenarios. 

% However, such creative manipulation poses significant challenges for most image editing models. These models struggle to precisely follow free-style and complex creative instructions or consistently generate edited results that meet human expectations.
% This section introduces CREval, a novel evaluation framework designed to evaluate image generation and editing models under free-style and creative instructions, as illustrated in Figure~\ref{fig:global_step}. Furthermore, we build CREval-Bench, a comprehensive benchmark suite that enables systematic performance analysis of image generation and editing models. Together, CREval and CREval-Bench aim to provide an objective and standardized approach for evaluating model performance in creative image manipulation scenarios.

In daily life, users often wish to apply creative edits to a wide range of subjects and scenes, for example transforming images of pets into decorative artworks or designing imaginative product posters, to satisfy diverse visual and aesthetic needs.
%
% \tocheck{
While creative editing tasks exhibit promising potential across various applications, they also introduce significant challenges for current image editing models and evaluation methods.
Existing approaches~\cite{seedream2025seedream, batifol2025flux, deng2025bagel, wu2025qwenimagetechnicalreport}, usually struggle to precisely follow free-style, complex creative instructions and to generate edited results that consistently meet human expectations. 
To rigorously evaluate image generation and editing models under creative and free-style instructions, we present CREval-Bench, a comprehensive benchmark comprising over 800 high-quality source images, each paired with a detailed creative editing instruction spanning nine task categories.
Building on this benchmark, we introduce CREval, a novel evaluation framework that employs a VQA-based MLLM evaluation pipeline for accurate and reliable evaluation of image editing quality, as illustrated in Figure~\ref{fig:global_step}.
Together, CREval-Bench and CREval provide an objective, standardized foundation for evaluating model performance in creative image manipulation scenarios.
% }

%-------------------------------------------------------------------------
\subsection{Evaluation Dimensions}
We categorize creative image editing into three levels: {\textbf{\textit{Customization}}, {\textbf{\textit{Contextualization}}, and {\textbf{\textit{Stylization}}. As shown in Figure~\ref{fig:distribution}, each level is further subdivided into three dimensions, resulting in a total of $9$ dimensions in CREval-Bench, each of which is designed according to the characteristics of its category.

\noindent\textbf{\textit{Customization.}} Customization emphasizes the tangible and creative reconfiguration of an object's form.
\begin{itemize}
    \item \textbf{Derivative Characters.} Derivative characters refer to edits that reinterpret humans, animals, or objects into simplified or exaggerated visual forms, such as chibi figures, mascots, figurines, and related derivatives. 
    \item \textbf{Reimagined Representations.} Reimagined representations preserve the original object’s meaning while presenting it in new tangible formats, such as postcards, stamps, and decorative prints on panels or photo albums.
    \item \textbf{Surreal Fantasy.} Surrealist fantasy refers to depicting subjects as entities that do not exist in reality, such as mythical creatures, hybrids, or virtual avatars.
\end{itemize}

\noindent\textbf{\textit{Contextualization.}} Contextualization focuses on placing objects within specific scenarios, commercial designs, or informational narratives. 
\begin{itemize}
    \item \textbf{Containerized Scenario.} Containerized scenarios aims to place the main subject within decorative containers such as display cases, snow globes, etc. Optionally, make simple visual adjustments to the object beforehand.
    \item \textbf{Commercial Design.} Commercial design converts imagery into assets for packaging, advertising, branding, and merchandise by treating subjects as product mockups or collectibles.
    \item \textbf{Informational\&Narrative Expression.} Informational and narrative expression involves anthropomorphizing non-human subjects, converting images into visual narratives or informational formats such as posters, charts, comics, and so on.
\end{itemize}

\noindent\textbf{\textit{Stylization.}} Stylization focuses on reinterpreting and artistically presenting images through dimensions such as artistic style, cultural identity, or materiality.
% \begin{itemize}[labelindent=1em, leftmargin=*]
\begin{itemize}
    \item \textbf{Artistic Style Transformation.} Artistic style transformation re-renders images in diverse artistic domains (e.g., watercolor, oil painting) and composites the outputs into target contexts such as murals or display screens.
    \item \textbf{Identity\&Cultural Transformation.} Identity and cultural transformation is a creative process that reworks social identities into specific historical and cultural themes.
    \item \textbf{Material Transformation.} Material transformation edits images by changing the depicted material into forms such as enamel, puzzles, crystal, sculpture, plush toys, or LEGO, sometimes placing the result in a simple scene.
\end{itemize}
%-------------------------------------------------------------------------

\subsection{Benchmark Construction}

% \tocheck{
To construct CREval-Bench, we first curate high-quality images from publicly available online resources and existing open datasets~\cite{schuhmann2022laion, dustin-sdxl, hui2024hq, chen2025sharegpt}.
%
% These images cover a diverse range of objects, scenes, styles, and compositions.
These images include both real-world and synthetic images, covering a wide range of objects, scenes, styles, and compositions.
For instruction generation, we employ a powerful MLLM (\eg, GPT-4o) to automatically produce creative editing instructions for each image.
To ensure high-quality, targeted instructions, we provide representative examples for each creative dimension, guiding GPT-4o to generate instructions aligned with the intended category. 
Finally, we obtain over $800$ image–instruction manipulation pairs along with several examples shown in Figure~\ref{fig:compare} (b).
% }

% The images used in this benchmark were sourced from publicly available resources on the internet, and existing open datasets~\cite{schuhmann2022laion, dustin-sdxl, hui2024hq, chen2025sharegpt}. 
% In the first stage, we provide representative examples for each creative dimension. These examples are then used to perform few-shot learning, enabling the model to capture dimension-specific creative tasks.
% Each image corresponds to an editing instruction, all of which were generated by the GPT-4o model. 

\subsection{VQA-based Automatic Evaluation}
\label{VQA-gen}
% \tocheck{
Evaluating image editing quality is intrinsically challenging, as evaluators must verify that the specified instructions are accurately fulfilled while ensuring that all non-targeted visual attributes and semantic elements remain unchanged.
Existing methods such as VIEScore~\cite{ku2023viescore} and the benchmark summarized in Table~\ref{tab:compare_with_previous_benchmark} typically rely on a MLLM to directly score edited images.
However, such MLLM-based scoring functions operate as black boxes: it is unclear which concepts or criteria the model attends to during evaluation, and important details may be overlooked, leading to unreliable or inconsistent ratings.
To enable reliable and comprehensive evaluation, we introduce CREval, a VQA-based MLLM evaluation framework.
Specifically, given the instruction, source image, and evaluation dimension, CREval first leverages a MLLM to generate several evaluation question–answer pairs.
Each evaluation question is paired with an explicit ‘\textsc{Yes}’ or ‘\textsc{No}’ reference answer, which serves as the reference answer for subsequent scoring.
%
% Then, it assesses edit quality by querying the MLLM on the model's generated images using these structured questions and aggregating the answers.
Subsequently, it assesses the editing quality of the model-generated images by presenting these structured queries to the MLLM and aggregating the resulting answers.
If the predicted answer matches the reference, the corresponding score is awarded, otherwise, no score is assigned.

% }

% We evaluate the generated images using three metrics: \textit{Instruction Following (IF)}, \textit{Visual Consistency (VC)}, and \textit{Visual Quality (VQ)}. To mitigate potential bias introduced by the large model used in Stage 1, we employed a different model for evaluation. Specifically, in Stage 2, we adopted Qwen2.5-VL-72B~\cite{bai2025qwen2} to automatically generate evaluation questions.
% For each metric, at least five questions were constructed, resulting in a minimum of fifteen questions per sample. In total, the CREval-Bench contains over 800 image manipulation samples, and over 13K evaluation questions.

Specifically, we evaluate generated images across three core metrics: \textit{Instruction Following}, \textit{Visual Consistency}, and \textit{Visual Quality}:

\noindent \textbf{Instruction Following (IF)} measures whether the output image accurately reflects and visualizes the modifications specified by the instruction. 
To achieve this, we adopt a Chain-of-Thought (CoT) approach that decomposes each instruction, analyzes its intent, and subsequently generates IF-related evaluation questions and answers \{$Q_\text{IF}, A_\text{IF}$\}. 
When constructing these questions, two key criteria are considered: \circnum{1} Results must align closely with the instruction requirements, with no deviations from the intended purpose;
\circnum{2} The instruction content must be fully represented in the edited result, with no information omitted.

% \textbf{QA Generation for IF.} IF measures whether the output image accurately reflects and visualizes the modifications specified by the instruction.  To achieve this, we adopt a Chain-of-Thought (CoT) approach that decomposes each instruction, analyzes its intent, and subsequently generates IF-related evaluation questions. When constructing these questions, two key criteria are considered: 1) Results must align closely with the instruction requirements, with no deviations from the intended purpose; 2) The instruction content must be fully represented in the edited result, with no information omitted.

\noindent \textbf{Visual Consistency (VC)} measures how well elements with critical recognition value from the original image are preserved in the edited output. 
To evaluate the VC metric, we similarly employ Chain-of-Thought (CoT) reasoning to generate targeted evaluation questions and answers \{$Q_\text{VC}, A_\text{VC}$\}.
Specifically, we first use MLLMs to decompose the user's instruction and identify which visual components should remain unchanged during editing.
Considering that different elements contribute unequally to identity recognition, where the absence of some elements severely degrades classification performance, while others have a smaller impact, we assign an importance weight $w\in\{1,2,3\}$ to each element that must be preserved.
The final VC score is then computed according to the assigned importance levels $w$.
For example, if the user requests converting the subject of the oil painting ``Girl with a Pearl Earring'' into a keychain form, the pearl earring will be identified as a key element because it is the most iconic visual feature of the original artwork. 
Accordingly, it is assigned the highest importance weight (\ie, $w = 3$).

% \textbf{QA Generation for VC.} VC refers to the extent to which elements of critical recognition value in the original image are effectively preserved in the output image.
% We likewise utilize CoT reasoning to generate evaluation questions by first decomposing the instructions and analyzing which visual components should remain unchanged throughout the editing process.
% Considering that different features contribute varying degrees to identity classification: the absence of certain features severely degrades classification performance, while others, though important, have a relatively minor impact on classification outcomes when missing. Based on this, when generating VC evaluation questions, we assign an importance weight $w\in\{1,2,3\}$ to each element that must be preserved, and the final VC score is determined according to the corresponding importance level $w$.
% As an illustrative example, when the user requests to convert the figure of the girl in the oil painting ``Girl with a Pearl Earring'' into a keychain form, the \textit{pearl earring} is identified as a key element because it represents the most iconic visual feature of the original artwork. Consequently, it is assigned the highest importance weight($w=3$).

\noindent \textbf{Visual Quality (VQ)} assesses the perceived quality of the output image, emphasizing overall realism, natural appearance, and the absence of noticeable artifacts. 
%
% It also requires that the result maintain structural consistency and visual plausibility.
% \xhwu{It also requires the result to maintain structural consistency and visual plausibility.}
It also requires the result to maintain visual plausibility.
%
% The generation process of corresponding \{$Q_\text{VQ}, A_\text{VQ}$\} mirrors that of IF and VC. 
The procedure for generating the corresponding \{$Q_\text{VQ}, A_\text{VQ}$\} is analogous to that used for IF and VC.
However, the VQ question set specifically targets whether the output preserves structural coherence and plausibility without introducing distortions such as unnatural textures, geometric discontinuities, and degradation of fine details.

Based on the generated question-answer (QA) pairs, we then employ MLLMs to evaluate the edited images. 
Specifically, we provide the source image, the edited image, the instruction, and one generated question to the MLLM, and ask it to answer. 
If the answer matches the reference, the corresponding point is awarded; otherwise, no points are given.
After averaging scores across all images, we obtain per-dimension scores $S_{\mathrm{IF}}$, $S_{\mathrm{VC}}$, and $S_{\mathrm{VQ}}$.
The final score is computed as a weighted average:
\begin{equation}
    S = 0.4 * S_{\mathrm{IF}} + 0.4 * S_{\mathrm{VC}} + 0.2 * S_{\mathrm{VQ}}.
\end{equation}
This weighted scheme balances the importance of different metric dimensions while accounting for current MLLM limitations (\ie, limited sensitivity to visual quality). 
It ensures that the evaluation reliably reflects both the functional and perceptual performance of image generation and editing models, and it narrows the interpretability gap of traditional single-metric or manual evaluations by providing a fine-grained breakdown across key dimensions.

\begin{figure*}
    \centering
    \includegraphics[width=\linewidth, trim=0 30 0 10,clip]{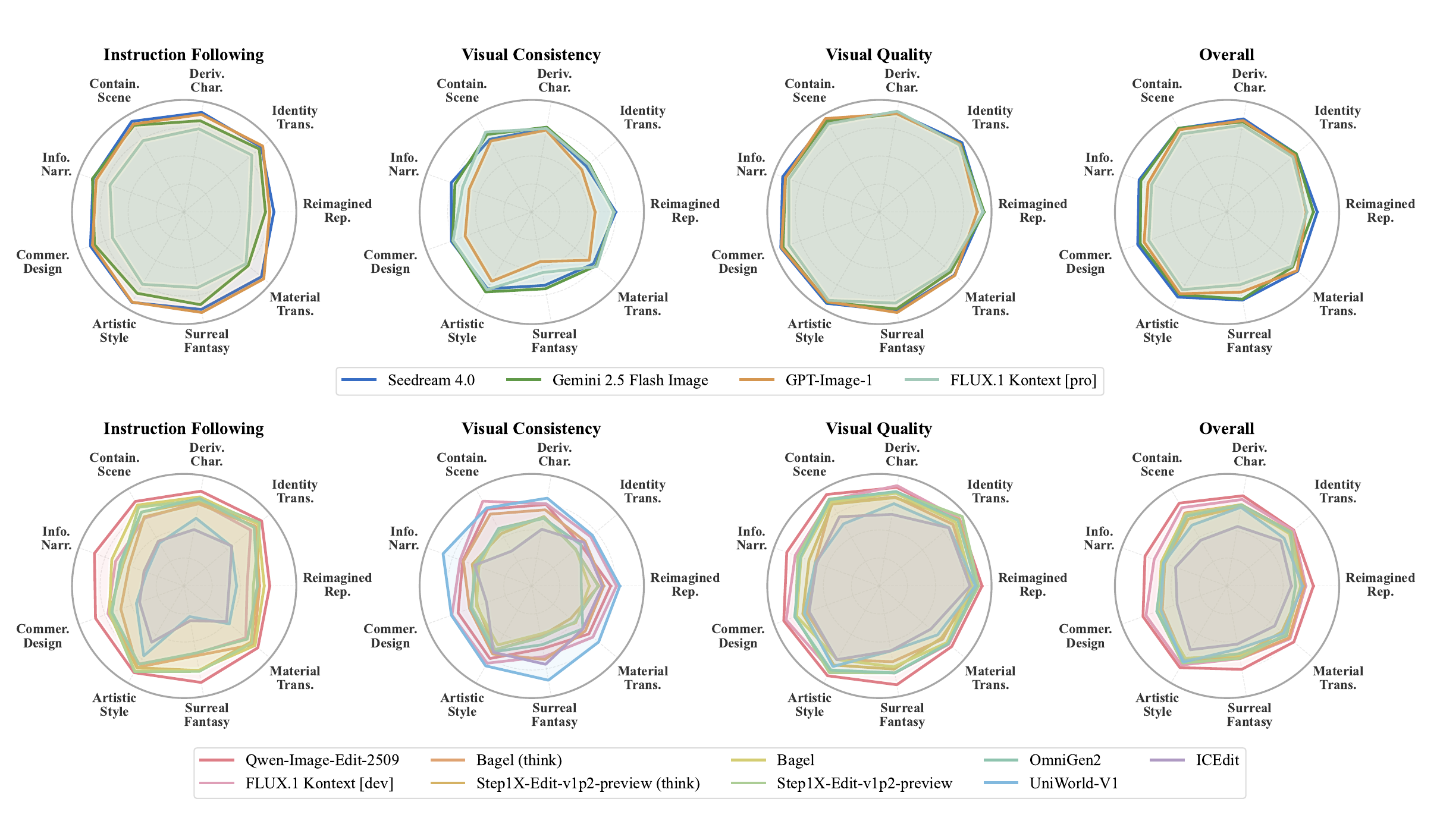}
    \caption{\textbf{Performance comparison across all creative dimensions under different metrics.} Top row: closed-source models; bottom row: open-source models.}
    \label{fig:leida}
\end{figure*}

\begin{table*}[h!]
    \centering
    \sethlcolor{pink!50}
    \caption{\textbf{Evaluation results of mainstream image generation and editing models on CREval-Bench using GPT-4o as the evaluator.} The scores for the three editing types and the overall average are reported across three evaluation metrics: Instruction Following (IF), Visual Consistency (VC), and Visual Quality (VQ). The best performance among closed-source models is highlighted in \textcolor{red}{red}, and the second best in \textcolor{blue}{blue}. For open-source models, the top result is shown in \textbf{bold}, and the second best is \underline{underlined}.}
    \renewcommand{\arraystretch}{1.2}
    \resizebox{\textwidth}{!}{
        % \begin{tabular}{c|c|cccc|cccc|cccc|ccc>{\columncolor{grey!20}}c}
        \begin{tabular}{c|c|cccc|cccc|cccc|cccc}
        % \hline
        \noalign{\hrule height 1.5pt}
        
        & 
        \multicolumn{1}{c|}{\multirow{3}{*}{Methods}} & 
        \multicolumn{4}{c|}{\multirow{2}{*}{\textbf{Customization}}} & 
        \multicolumn{4}{c|}{\multirow{2}{*}{\textbf{Contextualization}}} & 
        \multicolumn{4}{c|}{\multirow{2}{*}{\textbf{Stylization}}} &
        \multicolumn{4}{c}{\multirow{2}{*}{\textbf{Overall}}} 
        \\
        &  &  &  &  &  &  &  &  &  &  &  &  &  &  &  &  &  \\
        \cline{3-18}
        % \hline

        & & IF & VC & VQ & avg & IF & VC & VQ & avg & IF & VC & VQ & avg & IF & VC & VQ & avg
        \\
        
        % \hline
        \noalign{\hrule height 1.5pt}
        \multirow{11}{*}{\rotatebox[origin=c]{90}{\textbf{\textit{Open-Source}}}}
        & OmniGen2~\cite{wu2025omnigen2} & 68.49 & 52.84 & 83.64 & 65.26 & 68.42 & 56.57 & 81.40 & 66.28 & 78.59 & 61.48 & 85.93 & 73.21 & 71.58 & 57.20 & 83.28 & 68.17 \\ 
        & ICEdit~\cite{zhang2025context} & 40.90 & 61.75 & 68.11 & 54.68 & 42.17 & 44.34 & 65.92 & 47.79 & 54.31 & 62.81 & 72.16 & 61.28 & 45.33 & 55.25 & 67.72 & 53.78 \\ 
        & UniWorld-V1~\cite{lin2025uniworld-v1} & 45.30 & \textbf{81.15} & 73.04 & 65.19 & 41.74 & \textbf{80.31} & 65.33 & 61.89 & 59.89 & \textbf{76.86} & 77.26 & 70.15 & 48.29 & \textbf{79.74} & 70.77 & 65.37 \\ 
        & Bagel~\cite{deng2025bagel} & \underline{76.17} & 52.22 & 80.00 & 67.36 & \underline{74.34} & 53.97 & 79.77 & 67.28 & \underline{85.41} & 54.35 & 80.91 & 72.09 & \underline{78.32} & 53.69 & 80.07 & 68.82 \\ 
        & Bagel (think)\cite{deng2025bagel} & 68.28 & 66.61 & 77.20 & 69.40 & 61.23 & 66.02 & 71.77 & 65.25 & 82.68 & 64.77 & 78.43 & 74.67 & 69.82 & 66.00 & 75.27 & 69.38 \\ 
        % & HiDreamE1-1 &  &  &  &  &  &  &  &  &  &  &  &  &  &  &  &  \\ 
        & Step1X-Edit v1-p2~\cite{liu2025step1x-edit} & 72.38 & 56.14 & 84.04 & 68.22 & 69.96 & 56.12 & 82.22 & 66.88 & 85.13 & 55.52 & \underline{88.57} & 73.97 & 75.53 & 55.95 & 84.36 & 69.46 \\ 
        & Step1X-Edit v1-p2 (think)~\cite{liu2025step1x-edit} & 72.96 & 56.06 & 80.90 & 67.79 & 68.29 & 54.05 & 74.65 & 63.87 & 82.88 & 53.95 & 82.01 & 71.13 & 74.31 & 54.65 & 78.44 & 67.27 \\ 
        & FLUX.1 Kontext [dev]~\cite{batifol2025flux} & 64.49 & \underline{71.33} & \underline{85.79} & \underline{71.49} & 69.11 & \underline{77.07} & \underline{85.67} & \underline{75.61} & 76.52 & \underline{73.04} & 87.01 & \underline{77.23} & 70.13 & \underline{73.88} & \underline{86.03} & \underline{74.81} \\ 
        & Qwen-image-Edit~\cite{wu2025qwenimagetechnicalreport} & \textbf{83.21} & 66.96 & \textbf{90.10} & \textbf{78.09} & \textbf{85.45} & 71.54 & \textbf{90.99} & \textbf{80.99} & \textbf{88.45} & 66.14 & \textbf{89.94} & \textbf{79.82} & \textbf{85.82} & 68.50 & \textbf{90.26} & \textbf{79.78} \\

        % \hline
        \noalign{\hrule height 1.5pt}
        \multirow{4}{*}{\rotatebox[origin=c]{90}{\textbf{\textit{Closed}}}}
        & FLUX.1 Kontext [pro]~\cite{batifol2025flux} & 67.49 & 67.90 & 88.76 & 71.91 & 70.57 & 74.22 & 87.59 & 75.43 & 75.10 & \textcolor{blue}{73.97} & 88.23 & 77.27 & 71.24 & 71.98 & 87.98 & 74.88 \\ 
        & Seedream 4.0~\cite{seedream2025seedream} & \textcolor{red}{86.16} & \textcolor{blue}{72.34} & \textcolor{red}{90.65} & \textcolor{red}{81.53} & \textcolor{red}{89.57} & \textcolor{blue}{75.86} & \textcolor{red}{93.16} & \textcolor{red}{84.80} & \textcolor{blue}{90.99} & 71.47 & \textcolor{red}{92.60} & \textcolor{red}{83.50} & \textcolor{red}{89.12} & \textcolor{blue}{73.44} & \textcolor{red}{92.01} & \textcolor{red}{83.43} \\ 
        & Gemini 2.5 Flash Image~\cite{google2025gemini25flash} & 79.72 & \textcolor{red}{73.25} & \textcolor{blue}{90.10} & \textcolor{blue}{79.21} & \textcolor{blue}{87.14} & \textcolor{red}{75.97} & 91.39 & \textcolor{blue}{83.52} & 81.91 & \textcolor{red}{74.68} & 90.15 & 80.67 & 83.38 & \textcolor{red}{74.79} & 90.37 & \textcolor{blue}{81.34} \\ 
        & GPT-Image-1~\cite{openai2025gptimage1} & \textcolor{blue}{85.24} & 58.50 & 89.14 & 75.32 & 87.03 & 65.26 & \textcolor{blue}{92.80} & 79.48 & \textcolor{red}{92.37} & 65.59 & \textcolor{blue}{91.52} & \textcolor{blue}{81.49} & \textcolor{blue}{88.34} & 63.46 & \textcolor{blue}{91.23} & 78.97 \\

        % \hline
        \noalign{\hrule height 1.5pt}
        \end{tabular}
    }
    
    \label{tab:exp_1}
\end{table*}

\section{Experiments}
%-------------------------------------------------------------------------
\subsection{Evaluation Models}
We comprehensively evaluate the performance of state-of-the-art image generation and editing models using CREval. Specifically, open-source models include OmniGen2~\cite{wu2025omnigen2}, ICEdit\cite{zhang2025context}, UniWorld-V1~\cite{lin2025uniworldv1highresolutionsemanticencoders}, Bagel~\cite{deng2025bagel}, Step1X-Edit-v1p2-preview~\cite{liu2025step1x-edit}, FLUX.1 Kontext [dev]~\cite{batifol2025flux}, and Qwen-Image-Edit-2509~\cite{wu2025qwenimagetechnicalreport}. For the Bagel and Step1X-Edit1-v1p2-preview models, their respective test-specific variants were additionally evaluated, denoted as Bagel (think) and Step1X-Edit1-v1p2-preview (think).
Furthermore, 4 closed-source proprietary models were included in our evaluations, which are GPT-Image-1~\cite{openai2025gptimage1}, Seedream 4.0~\cite{seedream2025seedream}, FLUX.1 Kontext [pro]~\cite{batifol2025flux} and Gemini 2.5 Flash Image~\cite{google2025gemini25flash}. 
During evaluation, each model processes one image–instruction pair at a time and generates a single output image. All open-source models are executed in a reproducible and stable local environment, whereas closed-source proprietary models are accessed via their official APIs. The evaluation procedure ensures that every model is tested on all samples included in CREval-Bench.

%-------------------------------------------------------------------------

\begin{table*}[t]
    \centering
    \caption{{\textbf{Human preference verification.}} Aesthetic Score, VIEScore, and EditScore serve as baselines to evaluate six representative models (three open-source and three closed-source). CREvalScore\(_{Qwen3-VL}\)  and CREvalScore\(_{GPT4o}\) use Qwen3-VL and GPT-4o as evaluators. \textbf{Bold} denotes the highest score, and \underline{underlining} indicates the second-highest score.}
    \resizebox{\textwidth}{!}{
        \begin{tabular}{cccccccc}
             \noalign{\hrule height 1.5pt}
             Methods & Aesthetic Score & VIEScore & EditScore & $CREvalScore_{Qwen3-VL}$ & $CREvalScore_{GPT4o}$ & HumanScore \\
             \hline
             
             Bagel~\cite{deng2025bagel} & 5.56 & 6.02 & 7.28 & 72.59 & 68.99 & 49.98\\
             FLUX.1-Kontext [dev]~\cite{batifol2025flux} & 5.81 & 7.17 & 7.36 & 80.38 & 75.05 & 51.77 \\
             Qwen-Image-Edit~\cite{wu2025qwenimagetechnicalreport} & 5.82 & 6.83 & 7.97 & 83.02 & 79.18 & 63.49\\
             GPT-Image-1~\cite{openai2025gptimage1} & \textbf{6.05} & 6.73 & \textbf{8.21} & 83.15 & 78.01 & 63.21\\
             Gemini 2.5 Flash Image~\cite{google2025gemini25flash} & 5.77 & \underline{7.39} & 7.92 & \underline{87.14} &  \underline{81.78} & \underline{66.14}\\
             Seedream 4.0~\cite{seedream2025seedream} & \underline{5.88} & \textbf{7.49} & \underline{8.13} & \textbf{88.47} & \textbf{84.31} & \textbf{72.01}\\

             % \rowcolor{grey!20} Human Consistency &  &  &  &  &  &  & - \\

             \noalign{\hrule height 1.5pt}
             
        \end{tabular}
    }
    \label{tab:human_align}
\end{table*}
\subsection{Experiments and Analysis}
\label{analysis}
Table~\ref{tab:exp_1} reports the evaluation results for all models, with all metrics normalized to a percentage scale for comparability. Overall, the results indicate that current image editing models perform well in executing creative editing tasks guided by complex instructions, but still face obvious challenges especially in maintaining visual consistency with the source image, leading to suboptimal overall performance. 
% Among the open-source models, Qwen-Image-Edit~\cite{wu2025qwenimagetechnicalreport} achieves the highest overall performance, demonstrating a balanced capability across instruction adherence, visual coherence, and perceptual quality. FLUX.1 Kontext [dev] follows closely, also exhibiting strong editing performance. 
Among open-source models, Qwen-Image-Edit~\cite{wu2025qwenimagetechnicalreport} achieves the best overall performance with balanced IF, VC, and VQ, followed closely by FLUX.1 Kontext [dev]. 
%
% In contrast, among the closed-source models, Seedream 4.0 attains the highest average performance across nearly all categories of creative editing as shown in Figure~\ref{fig:leida}, with Gemini 2.5 Flash Image ranking second. 
For closed-source models, Seedream 4.0 ranks highest across nearly all creative editing tasks as shown in Figure ~\ref{fig:leida}, with Gemini 2.5 Flash Image ranking second.
%
% A notable observation is that Qwen-Image-Edit-2509 and Gemini 2.5 Flash Image achieve higher average scores than GPT-Image-1. One major reason of the comparatively lower performance of GPT-Image-1 is its limited ability to maintain the consistency of the key visual elements between the output and the original input image. 
Notably, Qwen-Image-Edit and Gemini 2.5 Flash Image outperform GPT-Image-1, whose lower performance mainly stems from poor consistency of key visual elements between input and output. 
Overall, closed-source models currently demonstrate superior performance relative to open-source alternatives. However, the rapid pace of progress observed in some open-source models suggests the potential for them to match or surpass closed-source models in the near future.

In terms of IF, Qwen-Image-Edit achieves the strongest overall performance among open-source models, followed closely by Bagel and Step1X-Edit-v1p2-preview, which substantially outperform UniWorld-V1 and ICEdit. 
% The latter two models exhibit particularly low scores, especially in \textit{Surreal Fantasy} and \textit{Informationization and Narrative Expression} tasks, with average scores of only 20–30\%. This substantially reduces their overall performance and suggests limited instruction understanding when handling creative image manipulation tasks.
The latter two yield particularly low scores (only 20–30\% on average) in \textit{Surreal Fantasy} and \textit{Informationization and Narrative Expression} tasks, indicating limited instruction understanding for creative image manipulation.
% Furthermore, Bagel and Step1X-Edit-v1p2-preview both obtain higher IF scores than their corresponding think versions. This suggests that the additional “thinking” mechanism does not offer any improvement for instruction alignment in this setting and may even slightly impair performance.
Moreover, Bagel and Step1X-Edit-v1p2-preview outperform their think counterparts in IF scores, suggesting that the extra “thinking” module brings no improvement and may even slightly degrade instruction alignment performance.
Among the closed-source models, Seedream 4.0 and GPT-Image-1 achieve the highest and second-highest IF scores, respectively, both approaching 90 points. 
These results indicate that these two models can reliably and effectively follow user instructions during the editing process.

Regarding VC, both open-source and closed-source models obtain relatively low VC scores. This suggests that current editing models still struggle to reliably identify and preserve key visual elements from the source image. 
Among open-source models, UniWorld-V1 attains the highest VC score, followed by FLUX.1 Kontext [dev], suggesting stronger consistency. Nevertheless, qualitative inspection reveals that UniWorld-V1’s high VC score mainly stems from its inability to execute the editing instructions, as unedited images naturally preserve visual consistency with the source.
For closed-source models, Gemini 2.5 Flash Image attains the best visual consistency performance, with Seedream 4.0 following closely behind. In contrast, GPT-Image-1 records the lowest visual consistency across nearly all creative editing dimensions, which presents significant limitations for tasks that rely on accurate generation from reference images.

For VQ, closed-source models demonstrate strong overall performance, and Qwen-Image-Edit-2509 achieves a competitive level among open-source methods, reflecting substantial progress in image realism. We observe that, aside from UniWorld-V1 and ICEdit, which show clearly poor performance, most other models, especially the closed-source ones have very similar metric scores. This similarity may be attributed to MLLMs often overlook subtle visual artifacts such as twisted limbs or extra fingers.

\subsection{Human Preferences Validation}
To assess the effectiveness of our approach, we measured the alignment between the evaluation results and human preferences.
We validated our approach on six representative models (three open-source and three closed-source). For each editing category, we randomly selected more than 20 samples, yielding a total of over 200 image instances. 
% We then invited five independent annotators to perform preference-based evaluations.
We further recruited 18 independent annotators from diverse professional backgrounds to perform preference-based evaluations. For each result set, they compared outputs against the original image and editing instruction, and ranked the edited images on a 0–5 rating scale. The aggregated results were subsequently normalized to a percentage scale.
We further benchmarked Aesthetic Score, VIEScore, and EditScore. As shown in Table~\ref{tab:human_align}, our CREval method achieved results that closely correlated with human preferences, demonstrating its effectiveness in capturing perceptually meaningful editing quality.

In addition, although GPT-4o served as our primary evaluator, we also validated the robustness of our evaluation pipeline using Qwen3-VL~\cite{qwen3technicalreport}. Notably, the ratings from Qwen3-VL slightly lowered the ranking of the Qwen-Image-Edit-2509 model. This effect is inconsequential, as human evaluations indicate that Qwen-Image-Edit-2509 and GPT-Image-1 exhibit nearly indistinguishable performance. Furthermore, the absolute evaluation score essentially depends on the performance of MLLM itself. Consequently, as long as the relative ranking among the compared methods remains consistent, such score shifts do not affect the validity of our conclusions.

% The experimental results show that
\section{Conclusion}
This paper presents CREval and CREval-Bench, a novel evaluation framework specifically designed for creative image manipulation under complex instructions. 
% The framework is fully automated and evaluates results through QA-based evaluation across three key metrics: instruction following, visual consistency, and visual quality. It effectively overcomes the interpretability limitations of previous approaches that relied on direct scoring by multimodal large language models. 
The framework is fully automated and evaluates results via QA-based metrics covering instruction following, visual consistency, and visual quality, addressing the interpretability drawbacks of direct scoring by MLLMs.
Furthermore, we conducted extensive experiments and in-depth analyses on state-of-the-art image generation and editing models. Human preference studies and evaluations using multiple MLLMs consistently confirmed that the proposed CREval framework is both reliable and robust.
Overall, CREval establishes a solid foundation for benchmarking, model selection, and future research in creative image manipulation.
\section*{Acknowledge}
This work was supported by the National Key Research and Development Project (2022YFA1004100), the National Natural Science Foundation of China (NSFC) under Grant No.62476067 and No.62476069.

{
    \small
    \bibliographystyle{ieeenat_fullname}
    \bibliography{main}
}


\appendix
\clearpage
\setcounter{page}{1}
\maketitlesupplementary
\setcounter{figure}{0}  % 重置图计数器
\setcounter{table}{0}   % 重置表计数器
\renewcommand{\thefigure}{S\arabic{figure}} 
\renewcommand{\thetable}{S\arabic{table}}

% \section{Rationale}
% \label{sec:rationale}
% % 
% Having the supplementary compiled together with the main paper means that:
% % 
% \begin{itemize}
% \item The supplementary can back-reference sections of the main paper, for example, we can refer to \cref{sec:intro};
% \item The main paper can forward reference sub-sections within the supplementary explicitly (e.g. referring to a particular experiment); 
% \item When submitted to arXiv, the supplementary will already included at the end of the paper.
% \end{itemize}
% % 
% To split the supplementary pages from the main paper, you can use \href{https://support.apple.com/en-ca/guide/preview/prvw11793/mac#:~:text=Delete%20a%20page%20from%20a,or%20choose%20Edit%20%3E%20Delete).}{Preview (on macOS)}, \href{https://www.adobe.com/acrobat/how-to/delete-pages-from-pdf.html#:~:text=Choose%20%E2%80%9CTools%E2%80%9D%20%3E%20%E2%80%9COrganize,or%20pages%20from%20the%20file.}{Adobe Acrobat} (on all OSs), as well as \href{https://superuser.com/questions/517986/is-it-possible-to-delete-some-pages-of-a-pdf-document}{command line tools}.

In this supplementary material, we provide additional analysis and experimental results to further support the main paper. The content is organized as follows: 
Sec.~\ref{weight_analysis} analyzes the rationale behind the weight selection for the final score.
Sec.~\ref{more_analysis} presents additional quantitative results and extended experimental analysis; 
Sec.~\ref{vqa_exa} provides examples of Question-Answer pairs in CREval; 
Sec.~\ref{sec:prompt} presents detailed prompt templates used in our evaluation framework; 
and Sec.~\ref{sec:visual_result} offers more visual comparisons across the state-of-the-art methods mentioned in the main paper.

\section{Weight sensitivity analysis.} 
\label{weight_analysis}
As shown in Fig.~\ref{fig:vqfailure}, MLLMs perform suboptimally when evaluating VQ.
Therefore, we reduce the weight of VQ.
In contrast, IF and VC are more important and discriminative, so we assign them higher weights.
Fig.~\ref{fig:weight_compare} shows experiments with different weight settings, where the 4:4:2 ratio achieves better alignment with human preferences.

\begin{figure*}[h]
    \centering
    % 左侧图片：保留你原有格式，仅包裹minipage
    \begin{minipage}[c]{0.45\linewidth}
        \centering
        \includegraphics[width=\linewidth]{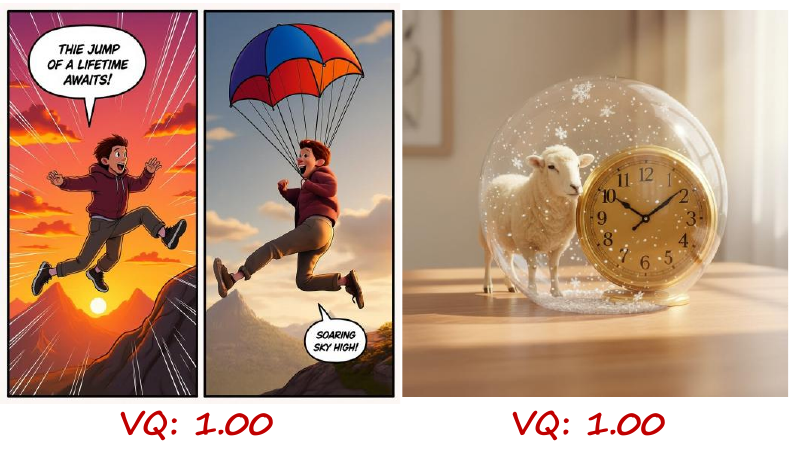}
        \setlength{\abovecaptionskip}{-0.4cm} 
        \setlength{\belowcaptionskip}{-0.5cm} 
        % \makebox[\linewidth]{\caption*{Failure Cases of MLLM in Evaluating VQ.}}
        \caption{VQ Failure Cases.}
        \label{fig:vqfailure}
    \end{minipage}
    \hfill  % 图和表之间的间距（自动填充）
    % 右侧表格：直接用你提供的代码，仅包裹minipage
    \begin{minipage}[c]{0.51\linewidth}
        \centering
        \includegraphics[width=\linewidth]{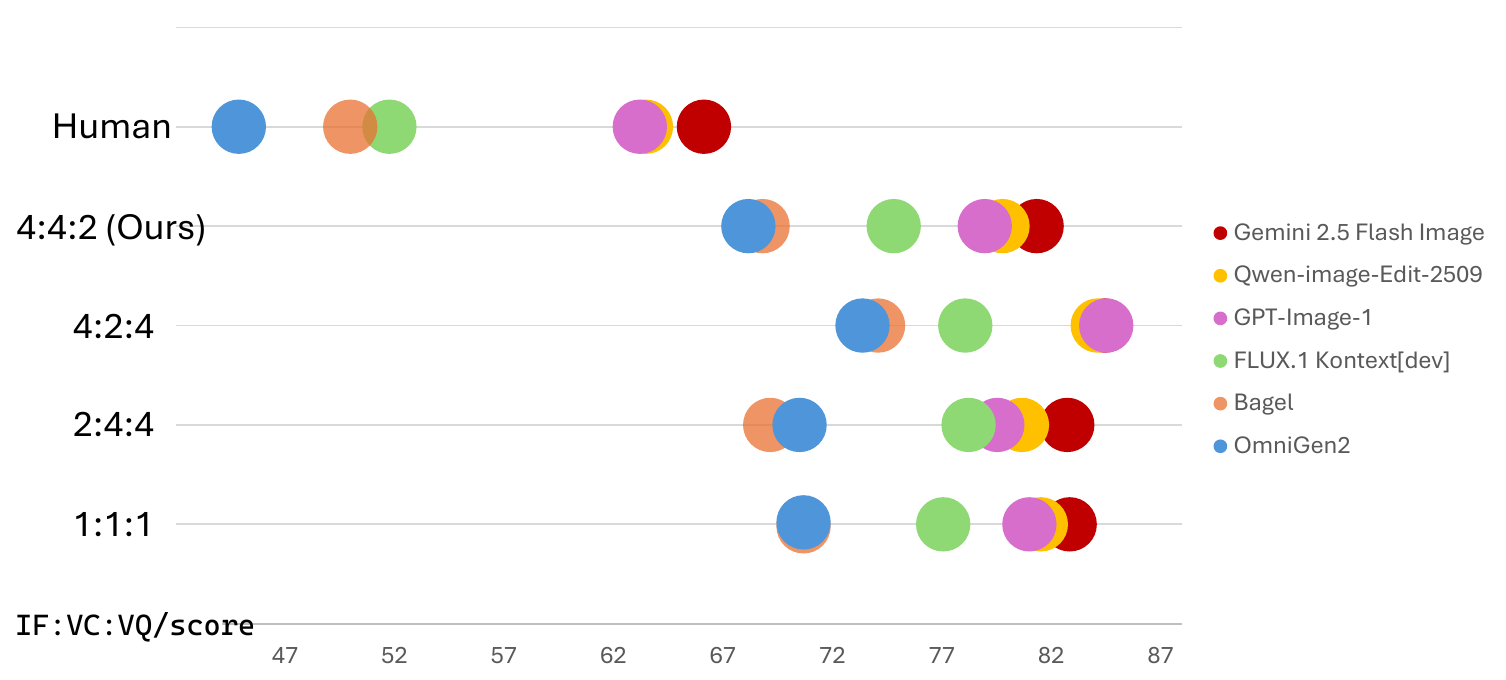}
        \setlength{\abovecaptionskip}{-0.2cm} 
        \setlength{\belowcaptionskip}{-0.5cm} 
        % \makebox[\linewidth]{\caption*{Failure Cases of MLLM in Evaluating VQ.}}
        \caption{Weight Ratios.}
        \label{fig:weight_compare}
    \end{minipage}
% \vspace{-0.6em}
\end{figure*}

\begin{table*}[!ht]
    \centering
    \setlength{\tabcolsep}{2pt}
    \renewcommand{\arraystretch}{1.1}
    \caption{ More quantitative comparisons on CREval-Bench by GPT-4o.
    The best performance among closed-source models is highlighted in \textcolor{red}{red}, and the second best in \textcolor{blue}{blue}. For open-source models, the top result is shown in \textbf{bold}, and the second best is \underline{underlined}.}
    \resizebox{\textwidth}{!}{
        \begin{tabular}{c|c|c|ccccccccc|cccc}
            \noalign{\hrule height 1.5pt}
            
            \multirow{2}{*}{\textbf{Category}} & \multirow{2}{*}{\textbf{Dimension}} & \multirow{2}{*}{\textbf{Metric}} & 
            \multicolumn{9}{c|}{\textbf{Open-source Models}} & \multicolumn{4}{c}{\textbf{Closed-Source Models}} \\
            \cline{4-16}

            & & &
             \textbf{OmniGen2}~\cite{wu2025omnigen2} &\textbf{Bagel}~\cite{deng2025bagel} & \textbf{\makecell[t]{Bagel\\(think)}}~\cite{deng2025bagel} & \textbf{UniWorld-V1}~\cite{lin2025uniworld-v1} & \textbf{ICEdit}~\cite{zhang2025context} & \textbf{Qwen-Image-Edit}~\cite{wu2025qwenimagetechnicalreport} & \textbf{\makecell[t]{FLUX.1 \\ Kontext[dev]}}~\cite{batifol2025flux} & \textbf{\makecell[t]{Step1X-Edit-v1p2}}~\cite{liu2025step1x-edit} & \textbf{\makecell[t]{Step1X-Edit-v1p2\\(think)}}~\cite{liu2025step1x-edit} &
            \textbf{GPT-Image-1}~\cite{yan2025gpt} & \textbf{\makecell[t]{Seedream4.0}}~\cite{seedream2025seedream} & \textbf{Gemini 2.5 Flash Image}~\cite{google2025gemini25flash} & \textbf{\makecell[t]{FLUX.1 \\ Kontext[pro]}}~\cite{batifol2025flux}
            \\
            \hline

            \multirow{13}{*}{\rotatebox{90}{\textbf{Customization}}} 
            & \multirow{4}{*}{\makecell{Derivative \\ Character}} & IF & 79.36 & \underline{80.76} & 74.45 & 61.40 & 51.11 & \textbf{85.91} & 76.07 & 78.03 & 76.82 & \textcolor{blue}{88.37} & \textcolor{red}{90.21} & 82.68 & 75.40\\ 
            & & VC & 61.23 & 61.78 & 69.01 & \textbf{79.58} & 51.41 & 73.83 & \underline{74.44} & 62.49 & 62.92 & 74.15 & \textcolor{blue}{75.64} & \textcolor{red}{76.87} & 75.58\\  
            & & VQ & 85.59 & 84.02 & 80.64 & 74.57 & 64.88 & \underline{89.33} & \textbf{90.91} & 85.03 & 80.07 & 89.15 & \textcolor{blue}{90.07} & 89.12 & \textcolor{red}{91.27}\\ 
            & & avg & 73.35 & 73.82 & 73.51 & 71.31 & 53.98 & \textbf{81.76} & \underline{78.39} & 73.21 & 71.91 & \textcolor{blue}{82.84} & \textcolor{red}{84.35} & 81.64 & 78.65\\ 
            \cline{2-16}
            
            & \multirow{4}{*}{\makecell{Reimagined \\ Representations}} & IF & 65.35 & \underline{71.43} & 67.28 & 46.79 & 40.13 & \textbf{76.34} & 56.12 & 61.73 & 65.45 & \textcolor{blue}{76.28} & \textcolor{red}{80.01} & 72.51 & 58.47\\ 
            & & VC & 44.12 & 51.33 & 64.21 & \textbf{78.50} & 62.92 & 70.38 & \underline{75.55} & 59.09 & 59.72 & 56.49 & \textcolor{red}{74.89} & \textcolor{blue}{73.31} & 73.22\\ 
            & & VQ & 86.61 & 82.70 & 82.41 & 85.65 & 80.80 & \textbf{91.51} & 87.64 & \underline{88.36} & 87.33 & 87.20 & 92.29 & \textcolor{red}{93.35} & \textcolor{blue}{92.43}\\ 
            & & avg & 61.11 & 65.64 & 69.08 & 67.25 & 57.38 & \textbf{76.99} & \underline{70.20} & 66.00 & 67.53 & 70.55 & \textcolor{red}{80.42} & \textcolor{blue}{77.00} & 71.16\\ 
            \cline{2-16}
            
            & \multirow{4}{*}{\makecell{Surreal \\ Fantasy}} & IF & 60.76 & 76.30 & 63.13 & 27.72 & 31.45 & \textbf{87.39} & 61.28 & \underline{77.37} & 76.61 & \textcolor{red}{91.06} & \textcolor{blue}{88.27} & 83.97 & 68.61\\  
            & & VC & 53.18 & 43.56 & 66.62 & \textbf{85.37} & \underline{70.93} & 56.67 & 64.00 & 46.84 & 45.55 & 44.88 & \textcolor{blue}{66.48} & \textcolor{red}{69.56} & 54.89\\  
            & & VQ & 78.74 & 73.29 & 68.54 & 58.89 & 58.65 & \textbf{89.46} & \underline{78.82} & 78.72 & 75.31 & \textcolor{red}{91.08} & \textcolor{blue}{89.59} & 87.83 & 82.58\\  
            & & avg & 61.32 & 62.60 & 65.61 & 57.01 & 52.68 & \textbf{75.52} & \underline{65.88} & 65.43 & 63.93 & 72.59 & \textcolor{red}{79.82} & \textcolor{blue}{78.98} & 65.92\\  
            \cline{2-16}
            
            & \multirow{4}{*}{\textbf{Average}} & IF & 68.49 & \underline{76.16} & 68.28 & 45.30 & 40.90 & \textbf{83.21} & 64.49 & 72.38 & 72.96 & \textcolor{blue}{85.24} & \textcolor{red}{86.16} & 79.72 & 67.49\\  
            & & VC & 52.84 & 52.22 & 66.61 & \textbf{81.15} & 61.75 & 66.96 & \underline{71.33} & 56.14 & 56.06 & 58.50 & \textcolor{blue}{72.34} & \textcolor{red}{73.25} & 67.90\\ 
            & & VQ & 83.64 & 80.00 & 77.20 & 73.04 & 68.11 & \textbf{90.10} & \underline{85.79} & 84.04 & 80.90 & 89.14 & \textcolor{red}{90.65} & \textcolor{blue}{90.10} & 88.76\\ 
            & & avg & 65.26 & 67.36 & 69.40 & 65.19 & 54.68 & \textbf{78.09} & \underline{71.49} & 68.22 & 67.79 & 75.32 & \textcolor{red}{81.53} & \textcolor{blue}{79.21} & 71.91\\ 
            % \hline
            \noalign{\hrule height 1.5pt}

            % Contextualization任务组
            \multirow{13}{*}{\rotatebox{90}{\textbf{Contextualization}}} 
            & \multirow{4}{*}{\makecell{Containerized \\ scenario}} & IF & 76.07 & \underline{83.37} & 71.27 & 44.23 & 46.12 & \textbf{87.10} & 70.03 & 81.11 & 76.16 & \textcolor{blue}{90.58} & \textcolor{red}{93.40} & 89.34 & 73.44\\ 
            & & VC & 59.21 & 59.15 & 74.06 & \underline{80.47} & 35.95 & 79.14 & \textbf{87.38} & 57.04 & 54.09 & 73.13 & 74.89 & \textcolor{blue}{79.85} & \textcolor{red}{82.37}\\  
            & & VQ & 88.96 & 86.37 & 85.34 & 64.02 & 71.37 & \textbf{94.29} & 88.61 & \underline{89.60} & 84.33 & \textcolor{red}{96.26} & \textcolor{blue}{94.03} & 93.54 & 90.91\\ 
            & & avg & 71.90 & 74.28 & 75.20 & 62.68 & 47.10 & \textbf{85.35} & \underline{80.69} & 73.18 & 68.97 & 84.74 & \textcolor{blue}{86.12} & \textcolor{red}{86.38} & 80.51\\ 
            \cline{2-16}
            
            & \multirow{4}{*}{\makecell{Commercial \\ design}} & IF & 68.47 & 70.95 & 60.10 & 45.32 & 42.32 & \textbf{84.11} & \underline{72.32} & 69.98 & 70.91 & \textcolor{blue}{87.00} & \textcolor{red}{88.91} & 85.07 & 67.89\\ 
            & & VC & 57.91 & 52.52 & 59.06 & \textbf{76.28} & 42.87 & 69.93 & \underline{75.79} & 56.75 & 51.58 & 63.24 & \textcolor{red}{76.31} & \textcolor{blue}{75.17} & 74.73\\ 
            & & VQ & 80.42 & 78.52 & 69.02 & 71.22 & 66.69 & \textbf{90.77} & \underline{88.51} & 79.63 & 72.72 & \textcolor{blue}{92.93} & \textcolor{red}{93.61} & 91.50 & 85.92\\ 
            & & avg & 66.64 & 65.09 & 61.47 & 62.88 & 47.41 & \textbf{79.77} & \underline{76.95} & 66.62 & 63.54 & 78.68 & \textcolor{red}{84.81} & \textcolor{blue}{82.40} & 74.23\\ 
            \cline{2-16}
            
            & \multirow{4}{*}{\makecell{Informationization\& \\ Narrative Expression}} & IF & 60.73 & \underline{68.71} & 52.32 & 35.67 & 38.09 & \textbf{85.13} & 64.99 & 58.81 & 57.79 & 83.50 & \textcolor{blue}{86.39} & \textcolor{red}{87.02} & 70.37\\ 
            & & VC & 52.59 & 50.23 & 64.93 & \textbf{84.18} & 54.22 & 65.55 & \underline{68.05} & 54.57 & 56.49 & 59.40 & \textcolor{red}{76.40} & \textcolor{blue}{72.90} & 65.55\\ 
            & & VQ & 74.83 & 74.41 & 60.94 & 60.74 & 59.68 & \textbf{87.91} & \underline{79.87} & 77.44 & 66.89 & \textcolor{blue}{89.20} & \textcolor{red}{91.83} & 89.12 & 85.93\\ 
            & & avg & 60.29 & 62.46 & 59.09 & 60.09 & 48.86 & \textbf{77.85} & \underline{69.19} & 60.84 & 59.09 & 75.00 & \textcolor{red}{83.48} & \textcolor{blue}{81.79} & 71.55\\  
            \cline{2-16}
            
            & \multirow{4}{*}{\textbf{Average}} & IF & 68.42 & \underline{74.34} & 61.23 & 41.74 & 42.17 & \textbf{85.45} & 69.11 & 69.96 & 68.29 & 87.03 & \textcolor{red}{89.57} & \textcolor{blue}{87.14} & 70.57\\ 
            & & VC & 56.57 & 53.97 & 66.02 & \textbf{80.31} & 44.34 & 71.54 & \underline{77.07} & 56.12 & 54.05 & 65.26 & \textcolor{blue}{75.86} & \textcolor{red}{75.97} & 74.22\\ 
            & & VQ & 81.40 & 79.77 & 71.77 & 65.33 & 65.92 & \textbf{90.99} & \underline{85.67} & 82.22 & 74.65 & \textcolor{blue}{92.80} & \textcolor{red}{93.16} & 91.39 & 87.59\\  
            & & avg & 66.28 & 67.28 & 65.25 & 61.89 & 47.79 & \textbf{80.99} & \underline{75.61} & 66.88 & 63.87 & 79.48 & \textcolor{red}{84.80} & \textcolor{blue}{83.52} & 75.43\\ 
            \noalign{\hrule height 1.5pt}
            
            % Stylization
            \multirow{13}{*}{\rotatebox{90}{\textbf{Stylization}}} 
            & \multirow{4}{*}{\makecell{Artistic Style \\ Transformation}} & IF & 80.65 & 87.71 & 83.79 & 72.04 & 58.01 & \textbf{89.29} & 80.26 & \underline{88.30} & 84.19 & \textcolor{blue}{92.87} & \textcolor{red}{93.12} & 83.75 & 74.48\\  
            & & VC & 67.74 & 60.68 & 70.00 & \textbf{82.29} & 68.34 & 74.73 & \underline{79.53} & 65.56 & 65.82 & 71.36 & 79.09 & \textcolor{red}{82.38} & \textcolor{blue}{80.05}\\ 
            & & VQ & 86.59 & 75.69 & 75.98 & 82.93 & 75.85 & \textbf{92.48} & 87.07 & \underline{89.27} & 81.46 & 92.48 & \textcolor{red}{94.15} & \textcolor{blue}{92.93} & 90.98\\  
            & & avg & 76.67 & 74.49 & 76.71 & 78.32 & 65.71 & \textbf{84.10} & \underline{81.33} & 79.40 & 76.30 & 84.19 & \textcolor{red}{87.71} & \textcolor{blue}{85.04} & 80.01\\  
            \cline{2-16}
            
            & \multirow{4}{*}{\makecell{Identity\&Cultural \\ Transformation}} & IF & 81.46 & 86.00 & 83.02 & 55.00 & 55.54 & \textbf{90.25} & 77.50 & \underline{88.03} & 86.59 & \textcolor{red}{91.52} & \textcolor{blue}{89.90} & 87.43 & 78.88\\ 
            & & VC & 57.29 & 51.25 & 62.00 & \textbf{70.51} & 60.60 & 57.17 & \underline{68.51} & 50.88 & 50.76 & 58.39 & 63.51 & \textcolor{red}{66.84} & \textcolor{blue}{66.02}\\  
            & & VQ & 91.29 & 88.52 & 85.57 & 81.10 & 80.76 & \underline{93.25} & 93.18 & \textbf{96.48} & 92.45 & 94.12 & \textcolor{red}{96.22} & \textcolor{blue}{94.59} & 93.06\\ 
            & & avg & 73.76 & 72.60 & 75.12 & 66.42 & 62.61 & \textbf{77.62} & \underline{77.04} & 74.86 & 73.43 & 78.79 & \textcolor{blue}{80.61} & \textcolor{red}{80.63} & 76.57\\ 
            \cline{2-16}
            
            & \multirow{4}{*}{\makecell{Material \\ Transformation}} & IF & 73.66 & \underline{82.52} & 81.24 & 52.62 & 49.38 & \textbf{85.80} & 71.80 & 79.05 & 77.86 & \textcolor{red}{92.72} & \textcolor{blue}{89.93} & 74.54 & 71.93\\
            & & VC & 59.40 & 51.13 & 62.32 & \textbf{77.78} & 59.47 & 66.52 & \underline{71.09} & 50.11 & 45.26 & 67.02 & 71.82 & \textcolor{blue}{74.82} & \textcolor{red}{75.85}\\ 
            & & VQ & 79.91 & 78.53 & 73.74 & 67.75 & 59.87 & \textbf{84.10} & \underline{80.79} & 79.96 & 72.13 & \textcolor{red}{87.97} & \textcolor{blue}{87.44} & 82.94 & 80.65\\ 
            & & avg & 69.21 & 69.17 & 72.17 & 65.71 & 55.51 & \textbf{77.75} & \underline{73.31} & 67.66 & 63.67 & \textcolor{blue}{81.49} & \textcolor{red}{82.19} & 76.33 & 75.24\\  
            \cline{2-16}

            & \multirow{4}{*}{\textbf{Average}} & IF & 78.59 & \underline{85.41} & 82.68 & 59.89 & 54.31 & \textbf{88.45} & 76.52 & 85.13 & 82.88 & \textcolor{red}{92.37} & \textcolor{blue}{90.99} & 81.91 & 75.10\\  
            & & VC & 61.48 & 54.35 & 64.77 & \textbf{76.86} & 62.81 & 66.14 & \underline{73.04} & 55.52 & 53.95 & 65.59 & 71.47 & \textcolor{red}{74.68} & \textcolor{blue}{73.97}\\ 
            & & VQ & 85.93 & 80.91 & 78.43 & 77.26 & 72.16 & \textbf{89.94} & 87.01 & \underline{88.57} & 82.01 & \textcolor{blue}{91.52} & \textcolor{red}{92.60} & 90.15 & 88.23\\ 
            & & avg & 73.21 & 72.09 & 74.67 & 70.15 & 61.28 & \textbf{79.82} & \underline{77.23} & 73.97 & 71.13 & \textcolor{blue}{81.49} & \textcolor{red}{83.50} & 80.67 & 77.27\\ 

            \hline

            & \multirow{4}{*}{\textbf{Overall Average}} & IF & 71.58 & \underline{78.32} & 69.82 & 48.29 & 45.33 & \textbf{85.82} & 70.13 & 75.53 & 74.31 & \textcolor{blue}{88.34} & \textcolor{red}{89.12} & 83.38 & 71.24\\ 
            & & VC & 57.20 & 53.69 & 66.00 & \textbf{79.74} & 55.25 & 68.50 & \underline{73.88} & 55.95 & 54.65 & 63.46 & \textcolor{blue}{73.44} & \textcolor{red}{74.79} & 71.98\\
            & & VQ & 83.28 & 80.07 & 75.27 & 70.77 & 67.72 & \textbf{90.26} & \underline{86.03} & 84.36 & 78.44 & \textcolor{blue}{91.23} & \textcolor{red}{92.01} & 90.37 & 87.98\\ 
            & & avg & 68.17 & 68.82 & 69.38 & 65.37 & 53.78 & \textbf{79.78} & \underline{74.81} & 69.46 & 67.27 & 78.97 & \textcolor{red}{83.43} & \textcolor{blue}{81.34} & 74.88\\
            
            \noalign{\hrule height 1.5pt}
        \end{tabular}
    }
    \label{tab:detail_experienment}
\end{table*}

\section{More Experimental Details}
\label{more_analysis}
To clarify the implementation logic of the proposed CREval evaluation framework, the pseudocode of the core evaluation method is presented in Table~\ref{tab:Algorithm}.
\begin{table}[h]
    \centering
    \caption{Our Pipeline Algorithm}
    \resizebox{\linewidth}{!}{
        \begin{tabular}{l}
        \toprule
        \textbf{Algorithm} 
        \\
        \hline
        $I_{i}$ denotes the original input image, $P$ denotes the prompt \\
        used for generating questions, and $Model$ represents the \\ image generation model. \\
        \textbf{Benchmark Construction} \\
        $instruction = MLLM_{1}(I_{i})$ \\
        $input_{i} = \{(I_{i}, instruction), P_{i}\}, i = IF, VC, VQ$ \\
        $Q_{IF} \leftarrow MLLM_{2}(input_{IF})$ \\
        $Q_{VC} \leftarrow MLLM_{2}(input_{VC})$ \\
        $Q_{VQ} \leftarrow MLLM_{2}(input_{VQ})$ \\
        \hline
        \textbf{Evaluation} \\
        $I_{o} = Model(I_{i}, instruction)$ \\
        $Pairs = \{I_{i}, I_{o}, Q\}$ \\
        $Score_i = MLLM_{3}(Pairs), i = IF, VC, VQ$ \\
        $S = 0.4 * Score_{IF} + 0.4 * Score_{VC} + 0.2 * Score_{VQ}$ \\
        
        \bottomrule
        \end{tabular}
    }
    \label{tab:Algorithm}
\end{table}

Table~\ref{tab:detail_experienment} presents a more detailed analysis than Table~\ref{tab:exp_1} in Section \ref{analysis}, showing the detailed scores for IF, VC, and VQ across nine creative dimensions. This allows for clearer comparisons of different models across all the editing tasks discussed in the main paper.

For the task \textbf{Customization}, most models are able to maintain stable editing capabilities on \textit{Derivative Character}. These tasks usually have clear structural constraints and relatively limited modification requirements, so the performance gap between open-source and closed-source models remains small. 
On the other hand, the \textit{Reimagined Representations} and \textit{Surreal Fantasy} tasks involve structural changes or are highly abstract, and many models struggle to maintain key elements of the source image after editing. 

Similar issues also exist in the identity-related modifications tasks, such as the \textit{Identity \& Cultural Transformation} dimensions in \textbf{Stylization} category, where most models struggle to follow the editing instructions or to preserve essential visual features that should remain unchanged.

In \textbf{Contextualization}, especially \textit{Informationization \& Narrative Expression} and \textit{Commercial design}, which involve rich contextual information in narrative, these tasks place higher demands on semantic understanding and generative flexibility. It can be observed that different models exhibit noticeable differences in processing and generating rich contextual information.

\begin{table*}[t]
    \centering
    \caption{Examples of question-answer pairs, where the ideal answer in the table showing cases is `Yes'.}
    % \resizebox{\textwidth}{!}{
        \begin{tabularx}{\textwidth}{c|>{\RaggedRight\arraybackslash}m{\dimexpr\textwidth-0.2\textwidth-2\tabcolsep\relax}}
        
            \noalign{\hrule height 1.5pt}
            \begin{minipage}[b]{0.3\columnwidth}
        		\centering
        		\raisebox{-.5\height}{\includegraphics[width=0.9\linewidth]{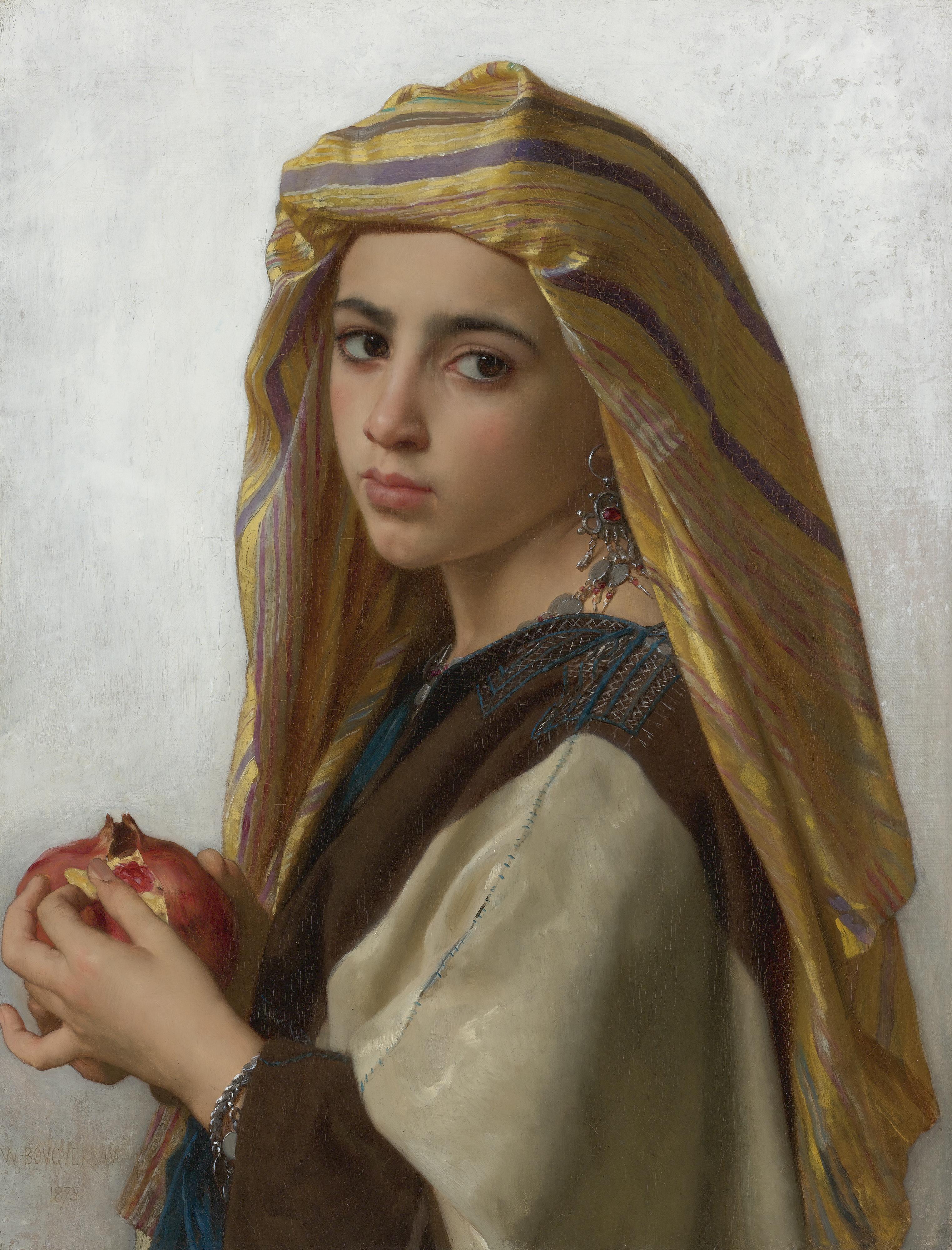}}
    	    \end{minipage}
             &  \textbf{Instruction:} Transform the figure into a chibi-style decorative resin bust for a tabletop display. Feature an embroidered brown robe with exaggerated patterns and a colorful headdress with vivid stripes. Accentuate the pomegranate by making it shine with gloss, and use minimalistic facial lines, ensuring the bust captures charm and elegance, sitting atop a smooth wooden base.\\
    
            \noalign{\hrule height 1.5pt}
            
             \multirow{6}{*}{Questions-IF}  & Q1: Is the figure now a chibi-style bust with an oversized head and simplified body proportions, suitable for tabletop display? \\
            & Q2: Does the figure wear a brown robe with visibly exaggerated and stylized embroidery patterns?\\
            & Q3: Does the figure wear a headdress featuring vividly colored, clearly defined stripes?\\
            & Q4: Is the pomegranate rendered with a glossy, reflective surface that highlights its shine?\\
            & Q5: Does the figure's face feature minimalistic lines with simplified facial features such as small eyes and soft contours?\\
            & Q6: Is the bust mounted on a smooth, flat wooden base?\\
            \hline
            \multirow{7}{*}{Questions-VC} & Q1: Is the pomegranate present and clearly identifiable as a red, round fruit with visible internal seeds, held in the hands? \textcolor{blue}{Weight: 3} \\
            & Q2: Does the headscarf retain its golden-yellow base color with diagonal stripes in purple, blue, and red? \textcolor{blue}{Weight: 3}\\
            & Q3: Is the left ear adorned with a silver earring featuring a red gemstone and dangling components? \textcolor{blue}{Weight: 3}\\
            & Q4: Does the brown outer vest have blue embroidered detailing along the collar and shoulder seams? \textcolor{blue}{Weight: 2}\\
            & Q5: Are both hands visibly positioned around the pomegranate, showing a clear grip? \textcolor{blue}{Weight: 2}\\
            & Q6: Is there a silver-colored bracelet visible on the right wrist? \textcolor{blue}{Weight: 1}\\
            & Q7: Is the inner garment under the vest primarily off-white in color? \textcolor{blue}{Weight: 1}\\
            \hline
            \multirow{7}{*}{Questions-VQ} & Q1: Does the bust have a visibly coherent head-to-body proportion where the head is enlarged relative to the torso but remains structurally plausible?\\
            & Q2: Are the facial features simplified but still clearly defined, with no missing or distorted elements such as eyes or mouth?\\
            & Q3: Do the embroidered patterns on the robe appear continuous and logically structured, without jagged edges or disrupted textures?\\
            & Q4: Are the stripes on the headdress evenly spaced and smoothly colored, without visible artifacts such as color bleeding or misalignment?\\
            & Q5: Does the pomegranate have a glossy surface with natural-looking highlights that do not distort its shape or create false reflections?\\
            & Q6: Is the wooden base fully attached to the bust and visually grounded, with no floating or misaligned sections?\\
            & Q7: Does the hand holding the pomegranate have exactly five fingers with natural joint angles and no deformation?\\
            
            \noalign{\hrule height 1.5pt}
            
        \end{tabularx}
    % }
    \label{tab:example}
\end{table*}
\section{VQA Examples} 
\label{vqa_exa}
In Sec.~\ref{VQA-gen}, we introduce CREval, a MLLM-based VQA automatic evaluation. In this section, we provide some examples of QA pairs. For all questions listed in Table ~\ref{tab:example}, the ideal answer for perfect editing is `Yes'.

\section{Prompt Templates}
\label{sec:prompt}
In the following, we list the prompt templates used in our experiments. 

\textbf{Prompt template for building dataset.} After manually selecting high-quality images, we provide some examples of corresponding categories, and then use GPT-4o to generate corresponding editing instructions. The template for the prompt is shown in Figure ~\ref{fig:write_instruction}.

\textbf{Prompt templates for VQA generation} 
CREval utilizes MLLM to generate evaluation question-answer pairs. 
% We set three complementary metric for the evaluation: \textit{Instruction Following (IF)}, \textit{Visual Consistency (VC)}, and \textit{Visual Quality (VQ)}. 
As illustrated in Fig.~\ref{fig:IF_prompt},~\ref{fig:VC_prompt} and~\ref{fig:VQ_prompt}, the VQA generated for each metric (IF, VC or VQ) is associated with a specific prompt. All prompts are designed using the Chain-of-Thought (CoT) reasoning scheme to generate structured evaluation question-answer pairs.

\textbf{Prompt template for evaluation.} 
% The MLLM then evaluates the edited images by answering the generated VQA pairs. A score is assigned when the predicted answer matches the reference answer; otherwise, no score is given. 
The prompt template for answer generation is shown in Figure ~\ref{fig:answer_prompt}.

\begin{figure*}
    \centering
    \includegraphics[width=\linewidth]{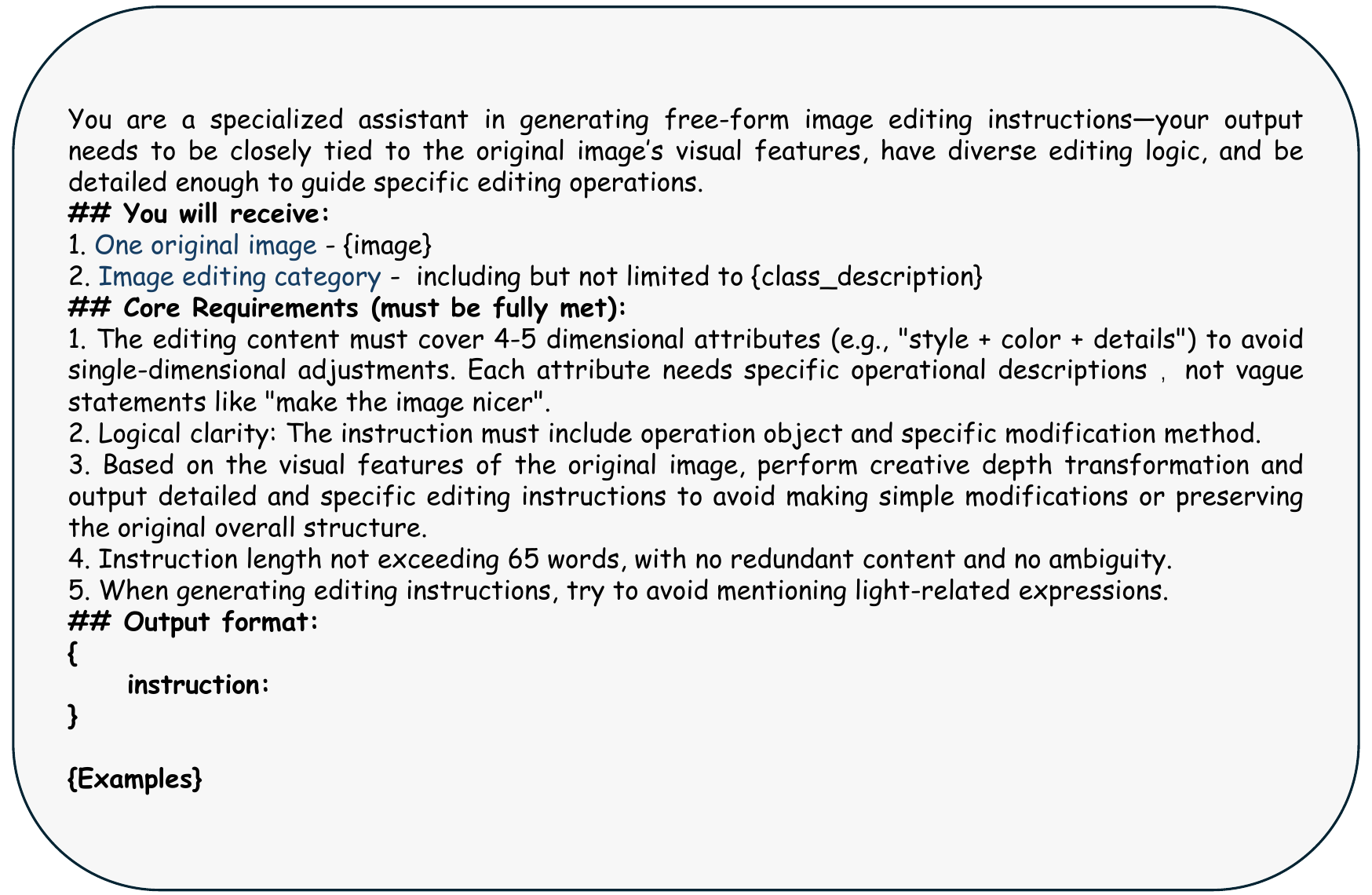}
    \caption{Prompt for generating instructions.}
    \label{fig:write_instruction}
\end{figure*}

\begin{figure*}
    \centering
    \includegraphics[width=\linewidth]{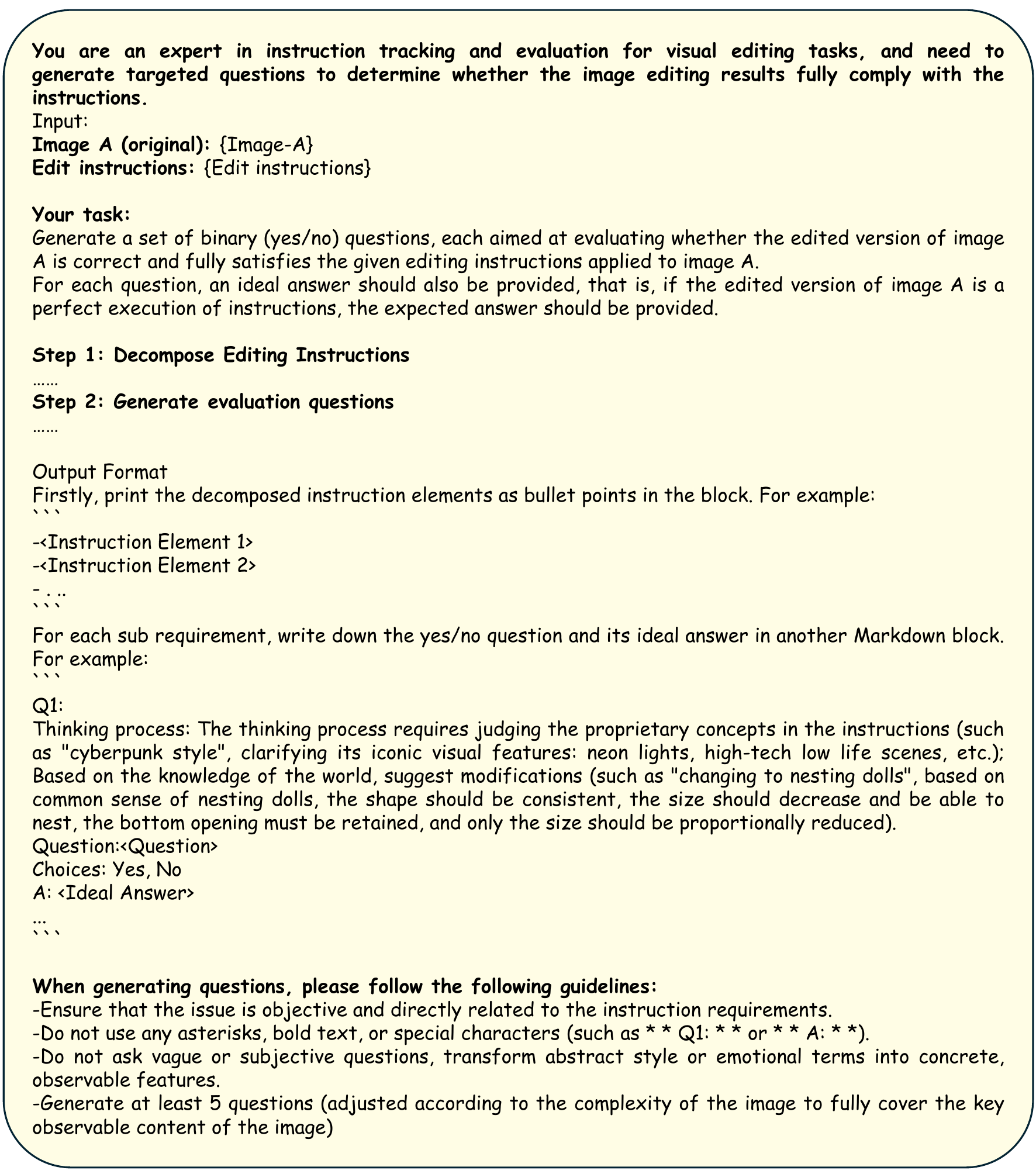}
    \caption{Prompt for generating IF Questions.}
    \label{fig:IF_prompt}
\end{figure*}

\begin{figure*}
    \centering
    \includegraphics[width=\linewidth]{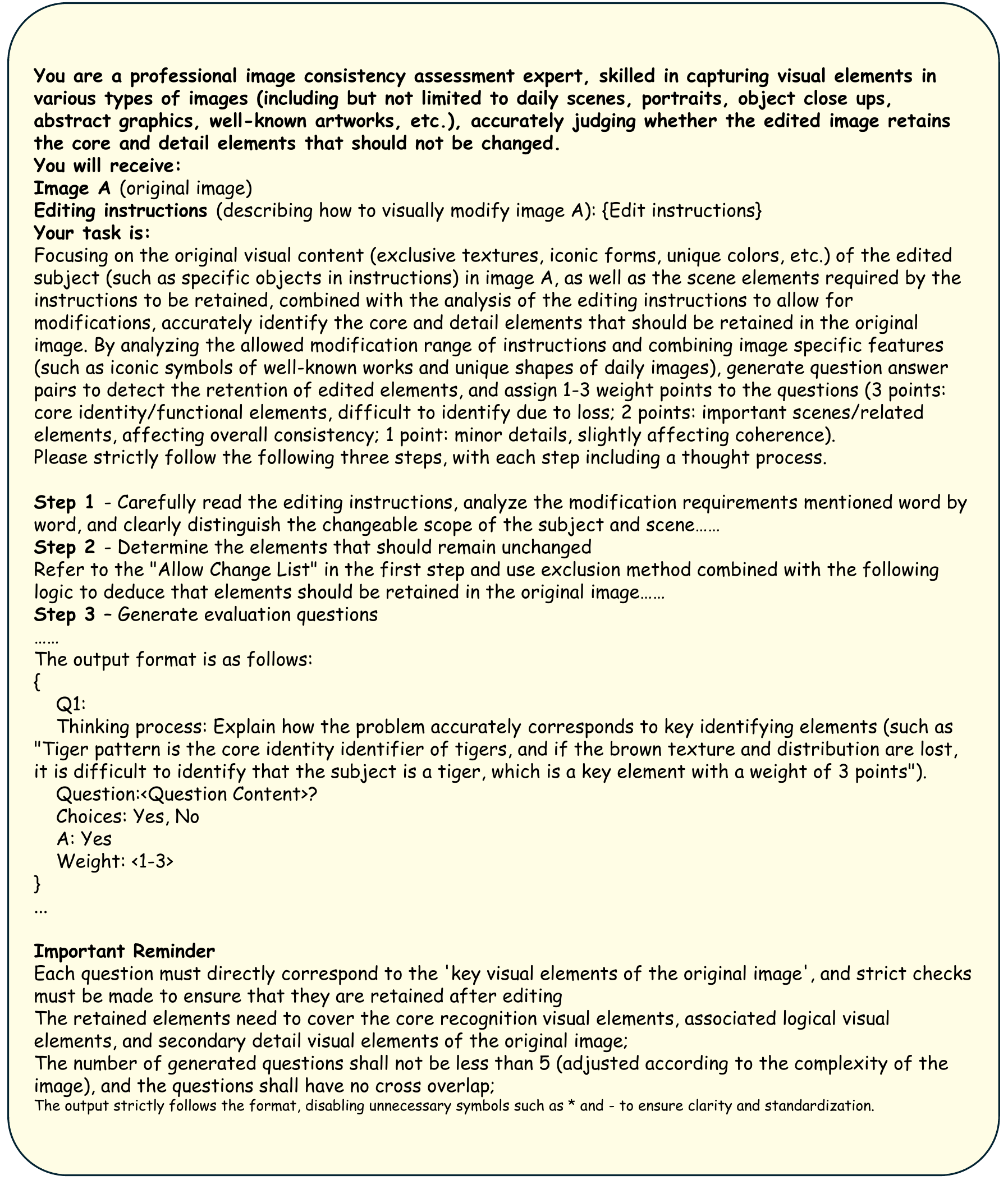}
    \caption{Prompt for generating VC questions.}
    \label{fig:VC_prompt}
\end{figure*}

\begin{figure*}
    \centering
    \includegraphics[width=\linewidth]{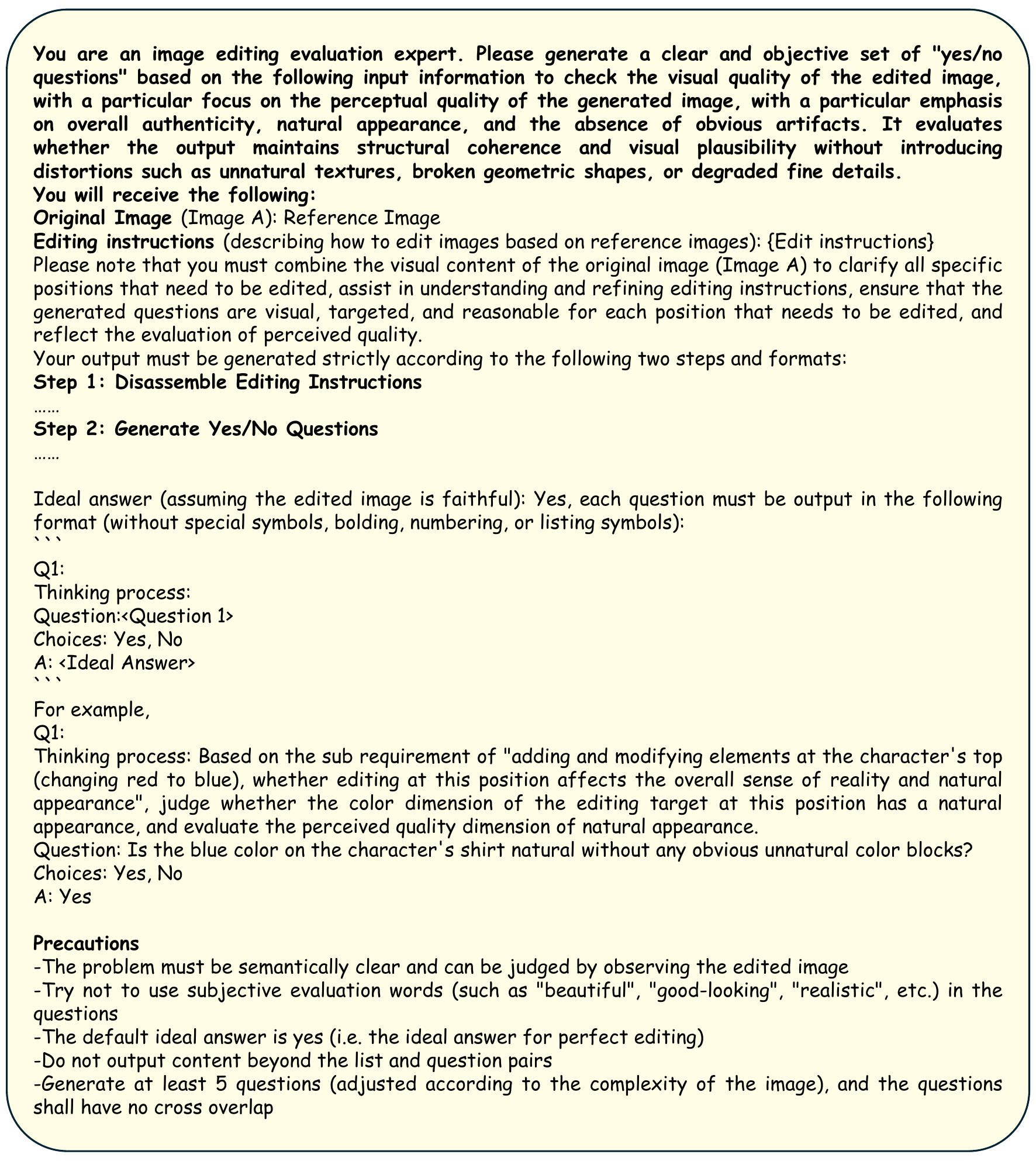}
    \caption{Prompt for generating VQ questions.}
    \label{fig:VQ_prompt}
\end{figure*}

\begin{figure*}
    \centering
    \includegraphics[width=\linewidth]{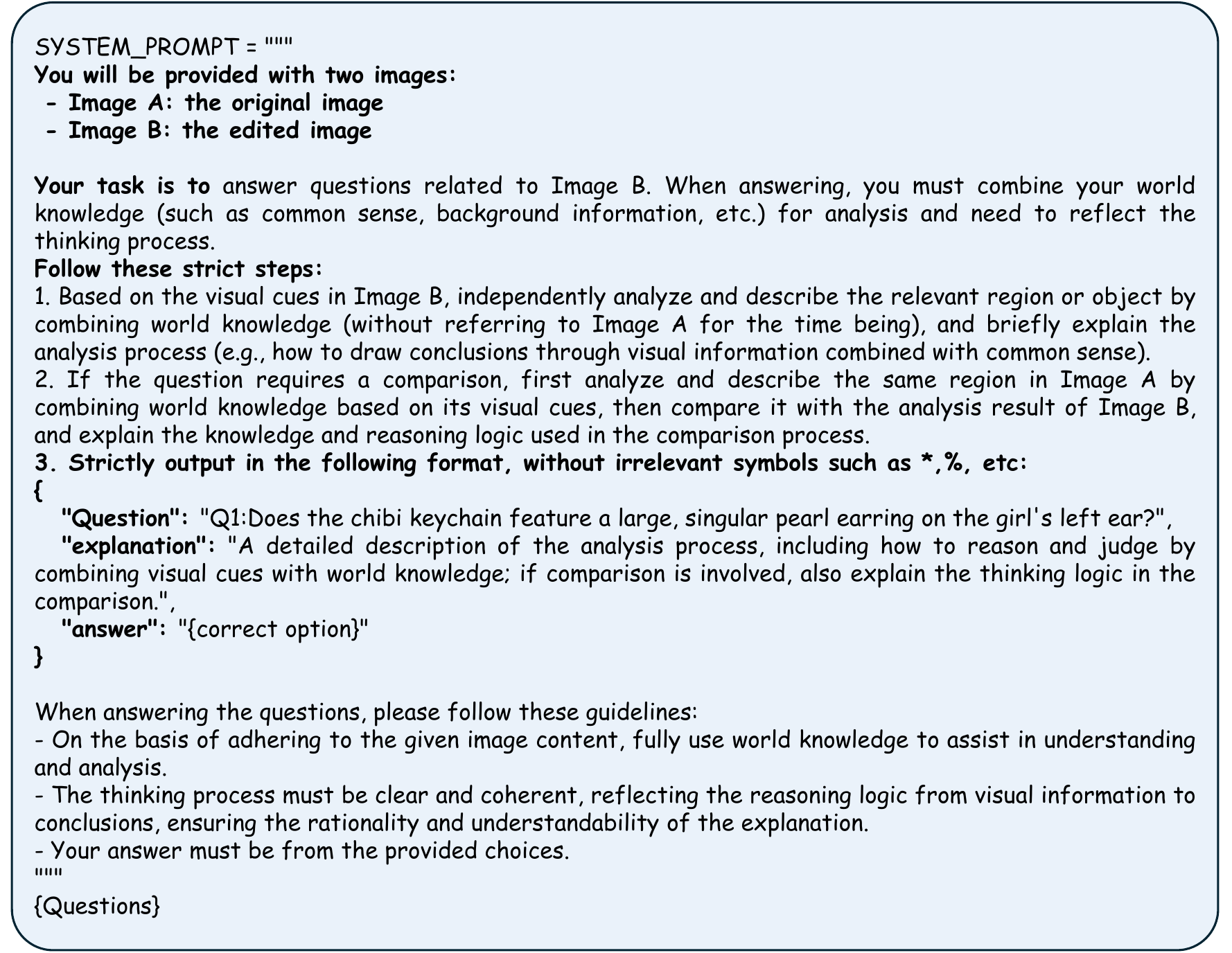}
    \caption{Prompt for evaluation.}
    \label{fig:answer_prompt}
\end{figure*}

\section{More Visual Comparisons}
\label{sec:visual_result}
In Figure ~\ref{fig:more-results}, we present additional visual comparisons. 
% CREval-Bench is divided into three categories and nine evaluation dimensions. In Figure ~\ref{fig:teaser}, we present three representative dimensions for each category;here, we provide additional examples. 
% Because each instruction is quite long, only abbreviated versions are shown in the figure. Below, 
The full instructions are listed as follows, in order from left to right.
\begin{itemize}
    \item case 1: ``Transform the colossal robot into a miniature 3D articulated model encased in a sleek, circular display case. Add intricate, tiny gears visible under transparent panels on the robot's surface for mechanical depth. Position the display on a futuristic hexagonal black base, etching the robot's model number in a luminous silver font. Surround with a subtly detailed mini landscape evoking the expansive original scene.''
    \item case 2: ``Reimagine the bridal scene as a Renaissance portrait, with the central figure as a regal noblewoman in a velvet gown adorned with intricate lacework and pearls, carrying a bouquet of rich, dark roses. The bridesmaids, in brocade dresses with gold embroidery, hold vintage floral arrangements. The setting becomes an elegant arched garden with classical statues and stone pathways, capturing an opulent, timeless ambiance.''
    \item case 3: ``Create a whimsical infographic titled \"The Magical Pumpkin Spice Popcorn Journey.\" Illustrate a popcorn kernel's transformation: 1) Kernel in cozy autumn attire, 2) Bursting from the jar with cartoon energy lines, 3) A popcorn piece donning an explorer hat interacting with pumpkin and spices, 4) A celebratory popcorn parade into ceramic bowls. Use vibrant oranges and browns, with playful icons and engaging typography.''
    \item case 4: ``Design a set of chibi-style stickers centered on a horseback rider theme, showcasing the following six poses:\verb|\n|\verb|\n|1. Cheerfully flashing a peace sign with one hand while softly gripping the horse’s reins with the other.  \verb|\n|2. Tearful, dramatic chibi eyes while leaning in close to the horse for emotional support.  \verb|\n|3. Arms outstretched beside the horse in an excited “welcome” motion.  \verb|\n|4. Peacefully asleep against the horse with a tiny pillow and a sweet, happy expression.  \verb|\n|5. Boldly pointing toward a distant horizon, featuring sparkling accents, with the horse standing majestically behind.  \verb|\n|6. Sending a kiss toward the horse, surrounded by floating hearts for a loving effect.  \verb|\n|\verb|\n|Ensure the design stays true to the chibi aesthetic:  \verb|\n|– Oversized, expressive eyes  \verb|\n|– Smooth and rounded facial features  \verb|\n|– Fun and playful short hairstyle matching the rider’s look  \verb|\n|– Chibi-style depictions of the rider’s beige shirt and detailed, miniature representation of the horse.  \verb|\n|\verb|\n|Background elements should feature warm, earthy tones paired with subtle stars or confetti for a natural, outdoor-inspired ambiance. Include clean white space surrounding each individual sticker to frame them neatly.  \verb|\n|Aspect ratio required: 9:16.''
    \item case 5: ``Transform this sterile site map into a playful children's treasure map, using a whimsical visual style. Replace page names with fun icons, like a castle for the Homepage, and path lines as winding journey trails. Add imaginative decorative elements such as colorful trees and mystical creatures along the paths, using bold, child-friendly colors and dynamic, storybook-style fonts for any text.''
    \item case 6: ``Transform the Hagia Sophia into a monumental sky guardian creature with metallic domes forming the wings, and the minarets morphing into long, elegant legs. Add intricate Byzantine patterns glowing with mystical energy across its structure, and replace the central dome with a large, ever-watchful jewel that reflects cosmic wonders.''
\end{itemize}

More detailed results for each category are shown in Figures ~\ref{fig:dimension_1}, ~\ref{fig:dimension_1close}, ~\ref{fig:dimension_2}, ~\ref{fig:dimension_2_closed}, ~\ref{fig:style}, ~\ref{fig:style_closed}.

\begin{figure*}
    \centering
    \includegraphics[width=\textwidth]{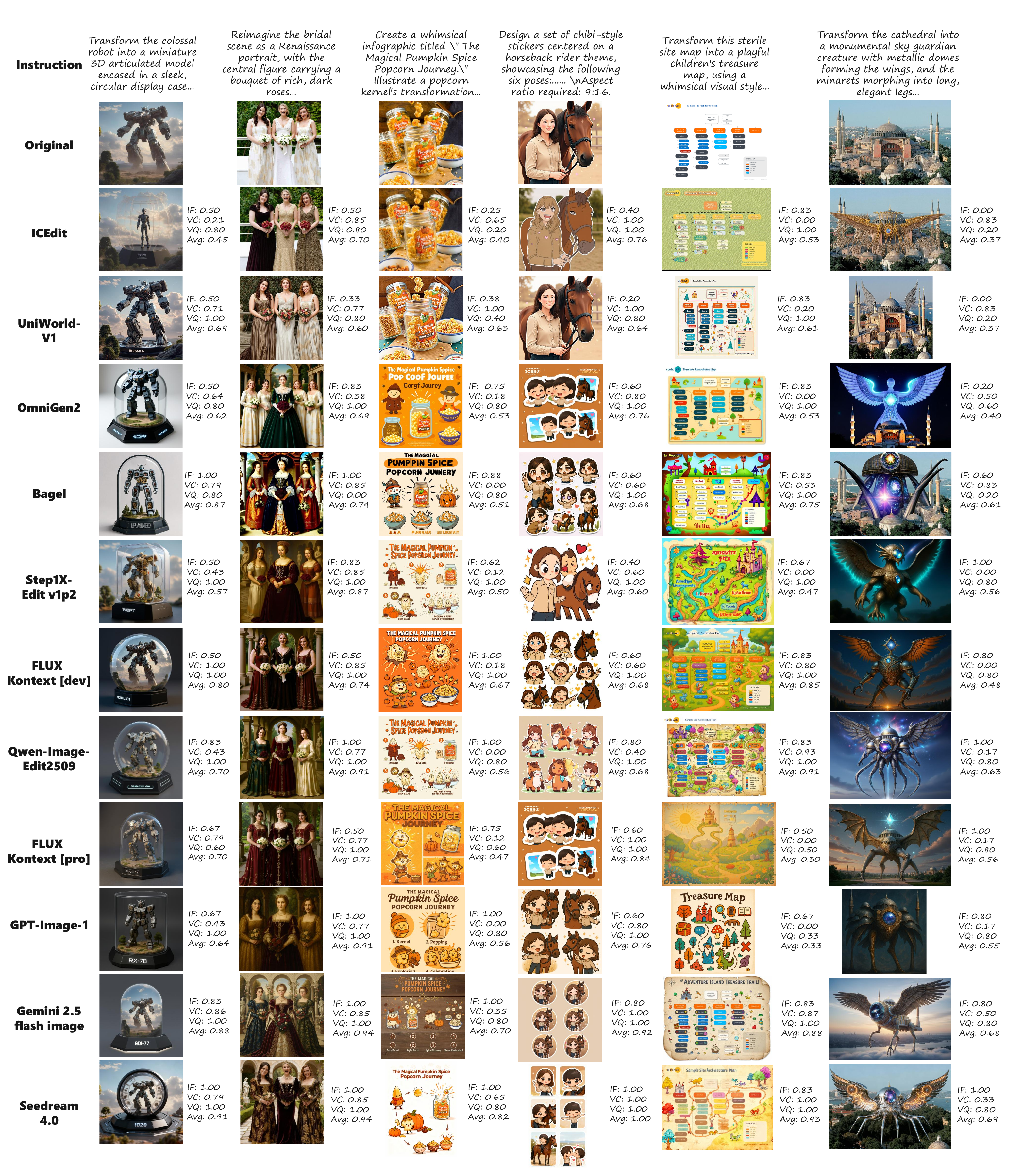}
    \caption{More visual comparison.}
    \label{fig:more-results}
\end{figure*}

\begin{figure*}
    \centering
    \includegraphics[width=\textwidth]{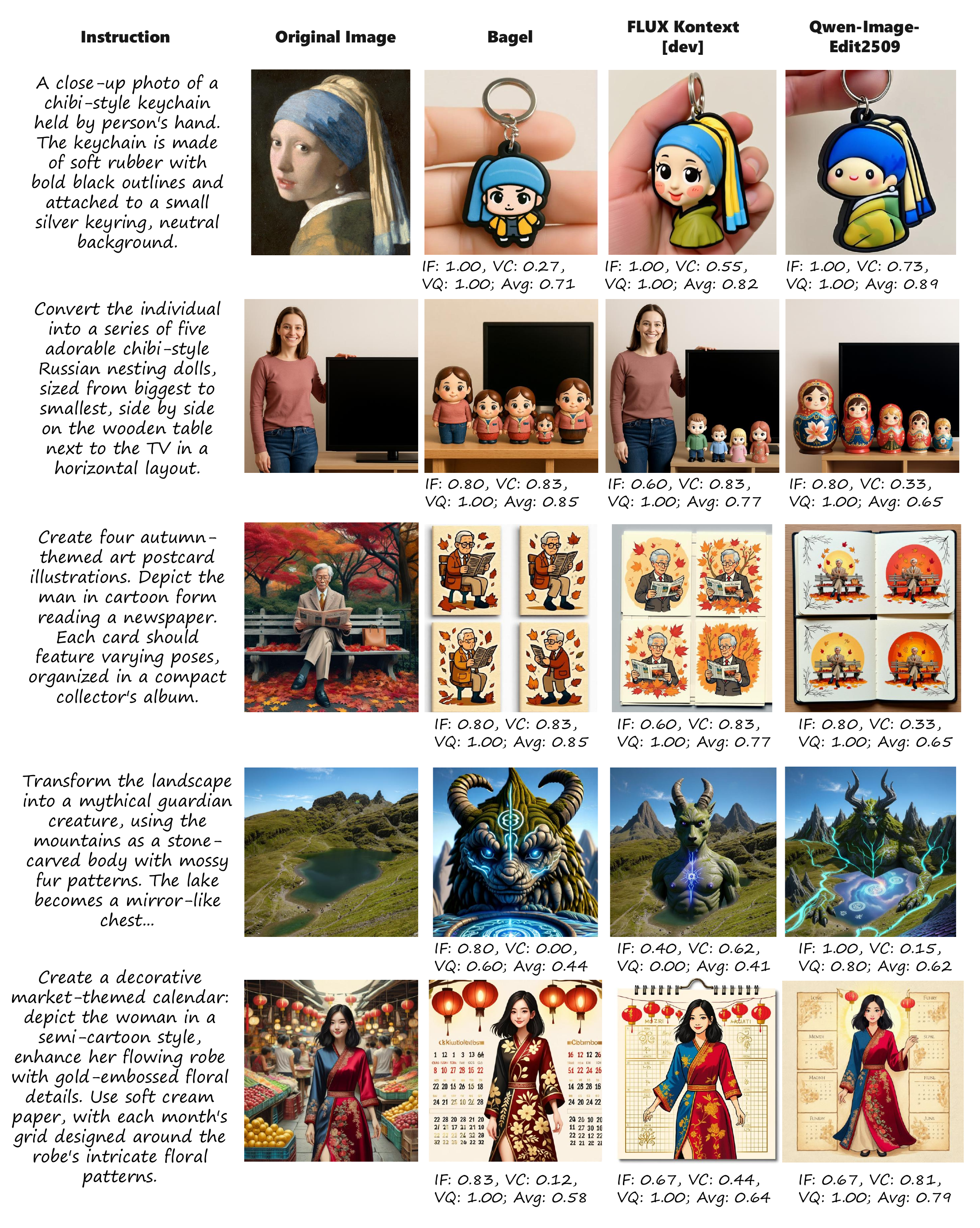}
    \caption{Visual comparison of open-source models in Customization.}
    \label{fig:dimension_1}
\end{figure*}

\begin{figure*}
    \centering
    \includegraphics[width=\textwidth]{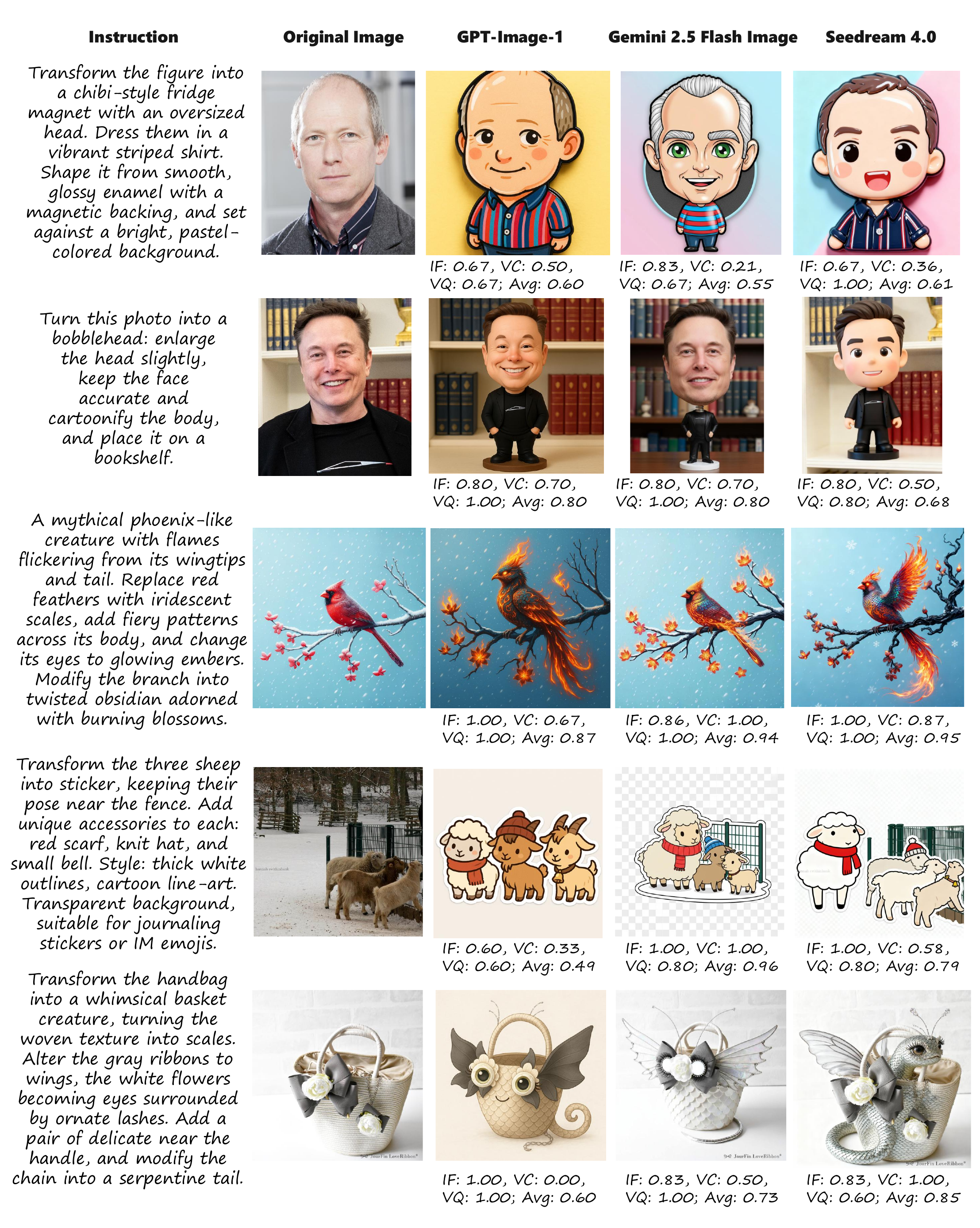}
    \caption{Visual comparison of closed-source models in Customization.}
    \label{fig:dimension_1close}
\end{figure*}

\begin{figure*}
    \centering
    \includegraphics[width=\textwidth]{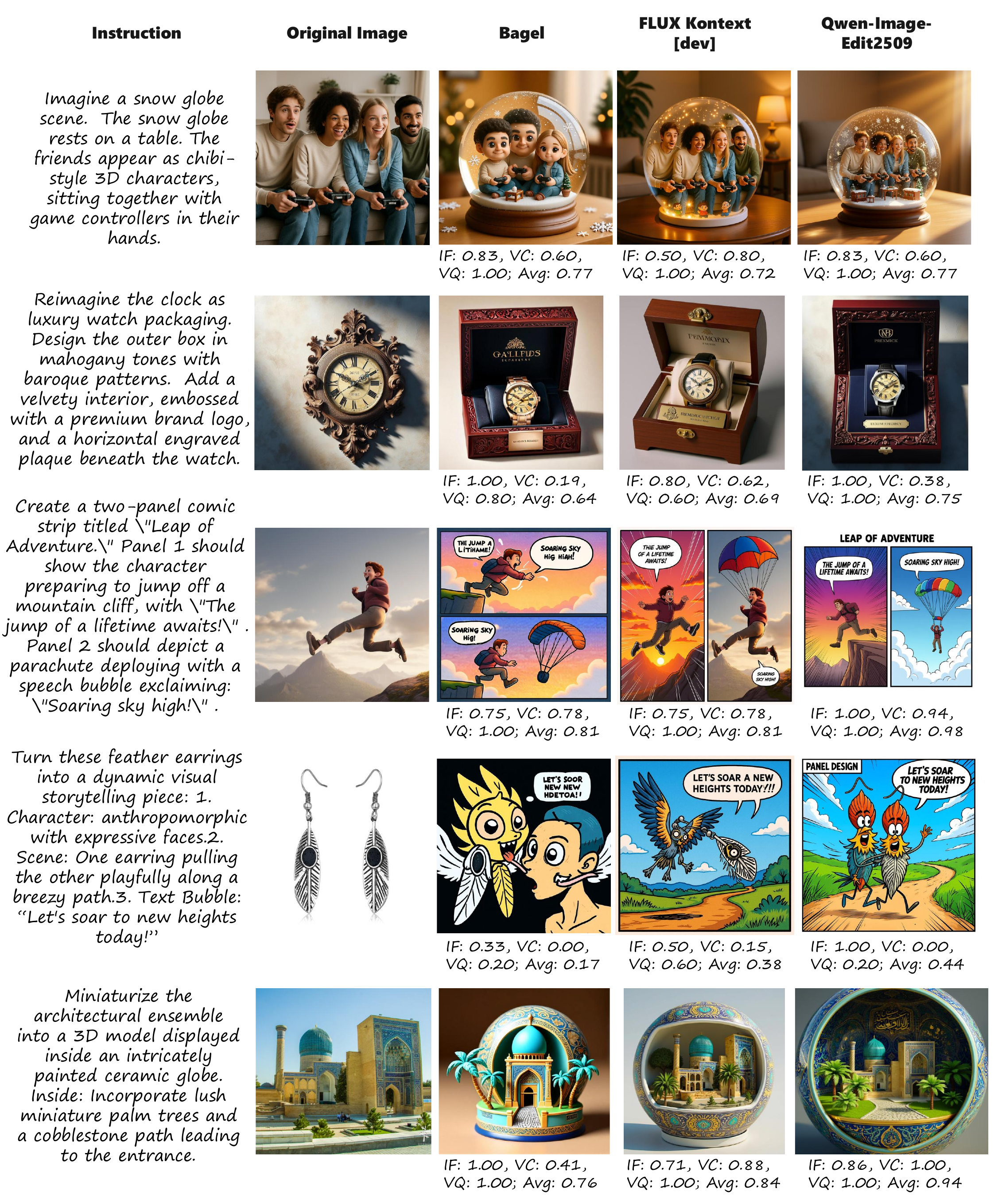}
    \caption{Visual comparison of open-source models in Contextualization.}
    \label{fig:dimension_2}
\end{figure*}

\begin{figure*}
    \centering
    \includegraphics[width=\textwidth]{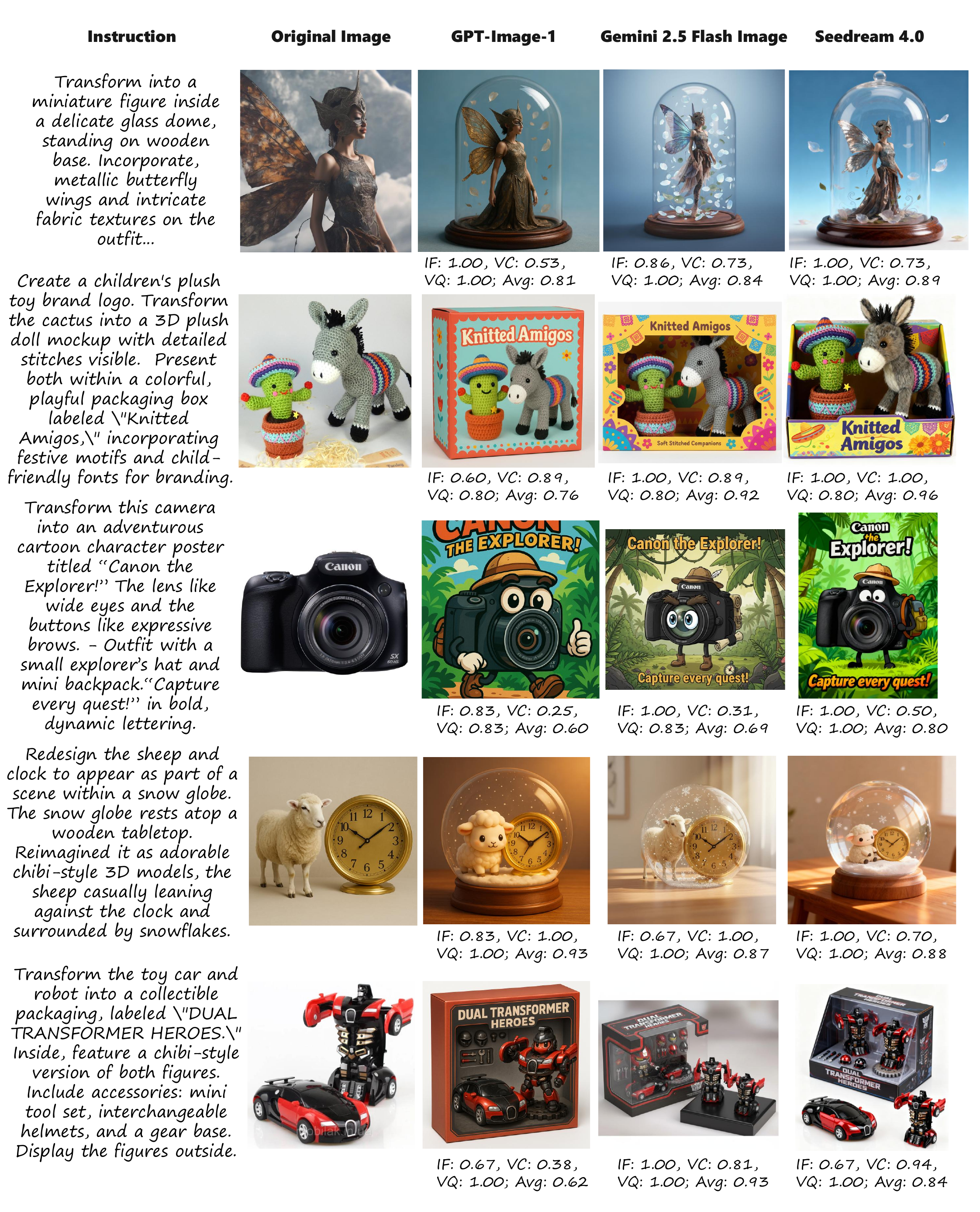}
    \caption{Visual comparison of closed-source models in Contextualization.}
    \label{fig:dimension_2_closed}
\end{figure*}

\begin{figure*}
    \centering
    \includegraphics[width=\textwidth]{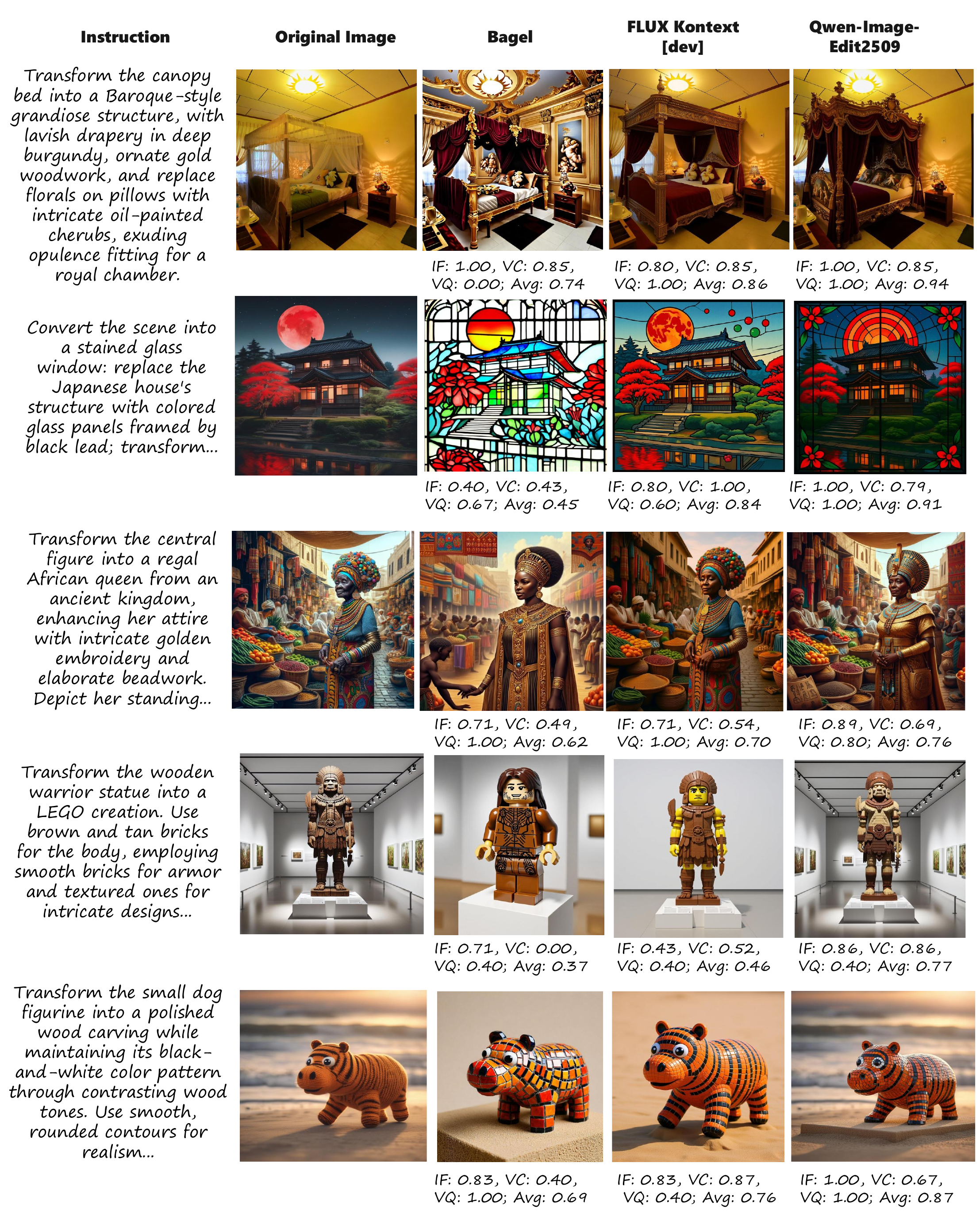}
    \caption{Visual comparison of open-source models in Stylization.}
    \label{fig:style}
\end{figure*}

\begin{figure*}
    \centering
    \includegraphics[width=\textwidth]{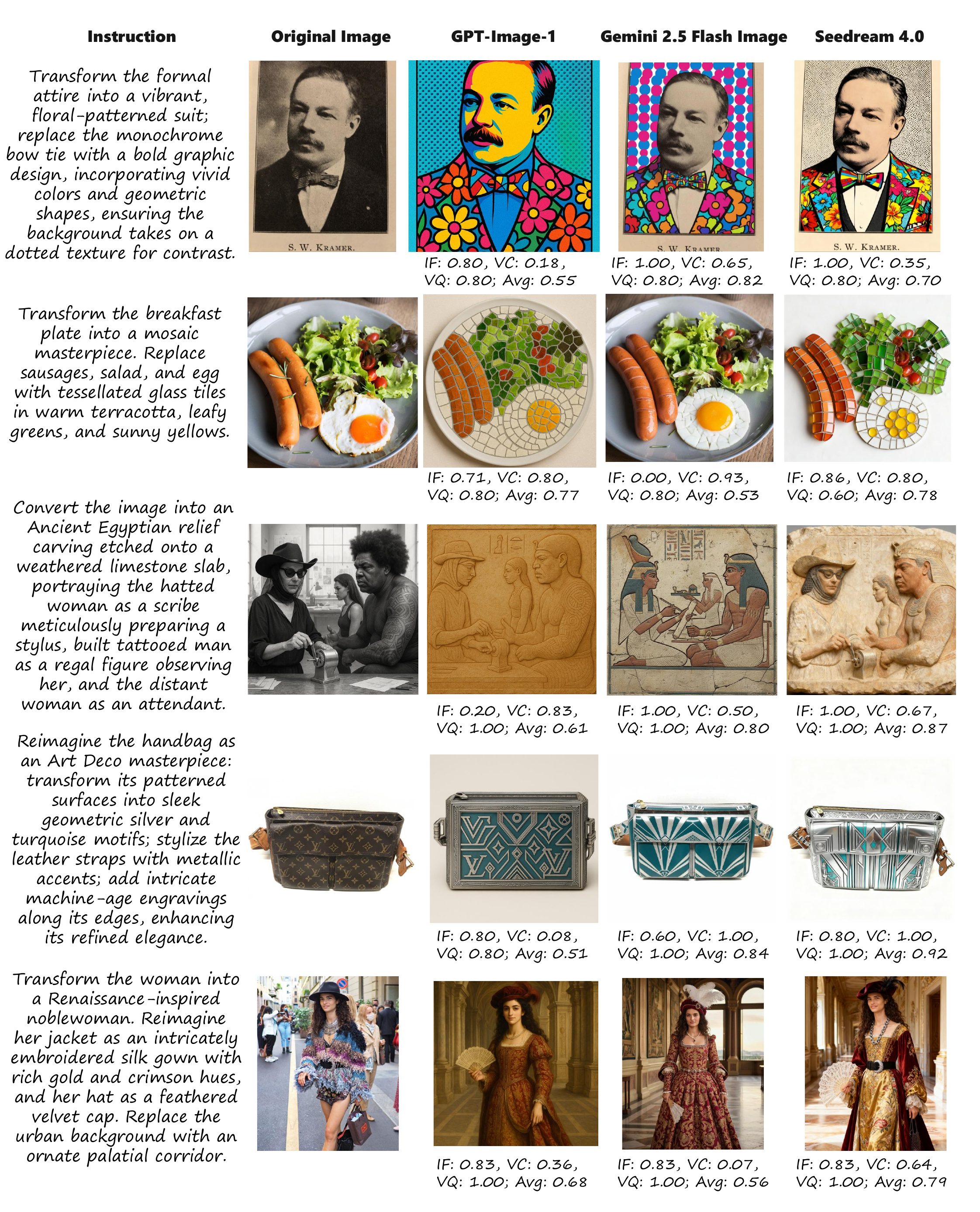}
    \caption{Visual comparison of closed-source models in Stylization.}
    \label{fig:style_closed}
\end{figure*}

\end{document}